%% file: main.tex
\documentclass[journal]{vgtc}

\onlineid{0}

\vgtccategory{Research}

\vgtcpapertype{please specify}

\title{ChartInsighter: An Approach for Mitigating Hallucination in Time-series Chart Summary Generation with A Benchmark Dataset}

\author{%
\authororcid{Fen Wang}{0009-0002-1039-2813},
\authororcid{Bomiao Wang}{0009-0007-4967-6414},
\authororcid{Xueli Shu}{0009-0003-5058-0541},
\authororcid{Zhen Liu}{0009-0009-1690-4745},
\authororcid{Zekai Shao}{0000-0003-2014-5293},
\authororcid{Chao Liu}{0000-0002-9748-3162}, and
\authororcid{Siming Chen}{0000-0002-2690-3588}
}

\authorfooter{
\item
Fen Wang is with Henan Institute of Advanced Technology, Zhengzhou University and Zhengzhou Zhongke Institute of Integrated Circuit and System Application. This work is done during visiting in Fudan University.
\item
Fen Wang, Bomiao Wang, Xueli Shu, Zhen Liu, Zekai Shao, Siming
Chen are with Fudan University. E-mail:  simingchen@fudan.edu.cn.
\item
Bomiao Wang and Xueli Shu contributed equally as second authors.\item
Chao Liu is with School of Computer Science, Fudan University and Zhengzhou Zhongke Institute of Integrated Circuit and System Application. E-mail: chaoliu20@fudan.edu.cn.
\item 
Chao Liu and Siming Chen are corresponding authors.
}

\abstract{
Effective chart summary can significantly reduce the time and effort decision makers spend interpreting charts, enabling precise and efficient communication of data insights. Previous studies have faced challenges in generating accurate and semantically rich summaries of time-series data charts. In this paper, we identify summary elements and common hallucination types in the generation of time-series chart summaries, which serve as our guidelines for automatic generation. We introduce ChartInsighter, which automatically generates chart summaries of time-series data, effectively reducing hallucinations in chart summary generation. Specifically, 
we assign multiple agents to generate the initial chart summary and collaborate iteratively, during which they invoke external data analysis modules to extract insights and compile them into a coherent summary. Additionally, we implement a self-consistency test method to validate and correct our summary. We create a high-quality benchmark of charts and summaries, with hallucination types annotated on a sentence-by-sentence basis, facilitating the evaluation of the effectiveness of reducing hallucinations. Our evaluations using our benchmark show that our method surpasses state-of-the-art models, and that our summary hallucination rate is the lowest, which effectively reduces various hallucinations and improves summary quality. The benchmark is available at 
\href{https://github.com/wangfen01/ChartInsighter}{https://github.com/wangfen01/ChartInsighter}.}


\usepackage{tabu}                      % only used for the table example
\usepackage{booktabs}                  % only used for the table example
\usepackage{lipsum}                    % used to generate placeholder text
\usepackage{mwe}                       % used to generate placeholder figures

\usepackage{mathptmx}                  % use matching math font

\begin{document}

%\firstsection{Introduction}

\maketitle
\input{sections/1-Introduction}

\input{sections/2-RelatedWork}
\input{sections/3-Preliminaries}

\input{sections/4-ChartInsighter}

\input{sections/5-Benchmark}
\input{sections/6-Evaluation}

\input{sections/7-Discussion}

\input{sections/8-Conclusion}

\section*{ACKNOWLEDGMENTS}

The authors want to thank the reviewers for their suggestions. This work is supported by Natural Science Foundation of China (NSFC No.62472099 and No.62202105).

\bibliographystyle{abbrv-doi-hyperref-narrow}

\bibliography{ref}

\appendix % You can use the hideappendix class option to skip everything after \appendix

\end{document}

%% file: sections/1-Introduction.tex
\section{Introduction}
 \textcolor{black}{Time-series data is widely present across various fields, including finance\cite{fu2011review}, energy\cite{singh2018big} and manufacturing\cite{wang2022detecting}. This widespread applicability makes time-series line charts one of the most commonly used visualization types on the Web\cite{battle2018beagle}. Automating the generation of time-series chart summaries is crucial for bridging the gap between raw data and data insights. It enables rapid comprehension, helping readers identify key insights\cite{kim2021towards} and improving recall and understanding of the data presented in charts\cite{hegarty1993constructing,large1995multimedia}.}

\begin{figure*}
\includegraphics[width=1\textwidth]{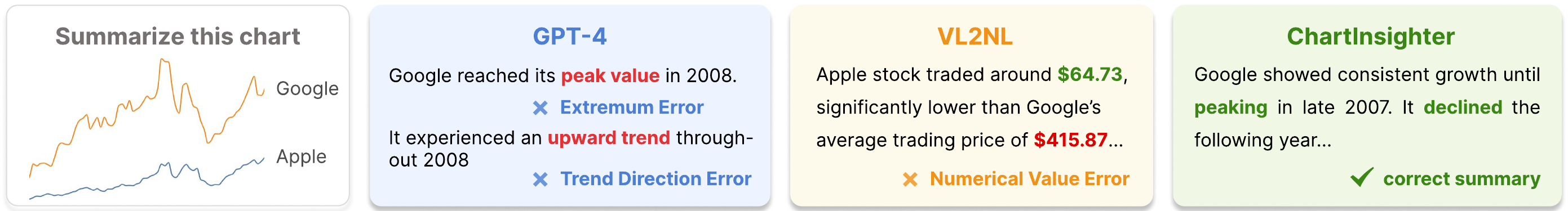}
\caption{\textcolor{black}{Examples of time-series chart summaries generated with GPT-4, VL2NL\cite{ko2024natural}, and ChartInsighter. Errors are indicated in red text, while correct points are highlighted in green text. GPT-4 makes an ``Extremum Error'', misidentifying 2008 as the peak year instead of the correct year, 2007, and a ``Trend Direction Error'', incorrectly describing a downward trend as an upward trend. VL2NL makes a ``Numerical Value Error'', incorrectly calculating Apple's average stock price. In contrast, ChartInsighter provides a correct summary.}}
\label{introduction}      
% Give a unique label
\vspace{-8px}
\end{figure*}

\textcolor{black}{Previous studies have utilized Large Language Models (LLMs) to automate the generation of chart summary, effectively enhancing the semantic richness of the summary\cite{ko2024natural,tang2023vistext,datatales}. 
However, time-series data, characterized by large volumes, high dimensionality, and complex variations, requires attention to specific data attributes, which may not be adequately captured by existing methods. These studies have primarily focused on basic chart analysis\cite{obeid2020chart,ko2024natural}, often overlooking a more in-depth exploration of trend analysis, data relationships, and detailed reasoning. Moreover, existing research\cite{tang2023vistext,obeid2020chart} frequently encounters hallucination issues, such as numerical calculation errors and incorrect trend judgments, which affect the accuracy and reliability of the generated summaries, as shown in Fig.~\ref{introduction}. Consequently, there remains a significant gap in research on generating chart summaries that can both provide a diverse and profound analysis of charts and mitigate hallucinations.}

Creating concise, accurate, and semantically rich time-series chart summaries with LLMs presents several challenges. 
Firstly, describing multidimensional time-series data requires capturing complex \textcolor{black}{relationships} across dimensions and over time. LLMs often lack deep \textcolor{black}{logical reasoning} about data context and the relationships between different data dimensions\cite{hadi2024large}. This deficiency can lead to various hallucinations in handling time-series data, significantly affecting the accuracy and reliability of chart summary.
Secondly, analyzing time-series data necessitates mathematical computations to identify data features and trends to recognize patterns within the data, where LLMs often fall short\cite{xu2024exploring}. \textcolor{black}{This limitation makes it even more challenging to extract valuable insights from time-series data.
Thirdly, LLMs organize statistical indicators such as mean and growth rate which help to better express the trends and characteristics of the data in summary, however}, the resulting semantics are often isolated and disjointed, lacking the smooth logical connections needed to form paragraphs with complete and fluid semantic flow.
These challenges highlight the need for enhanced capabilities in LLMs to accurately generate chart summaries that are not only precise but also contextually and semantically enriched. \textcolor{black}{ At the same time, the accuracy and reliability of summary evaluation have been a long-standing issue. Existing evaluation methods primarily focus on the semantic richness of the generated summaries\cite{tang2023vistext} and the similarity between generated summary and gold summary\cite{liu2024chartthinker}. However, the existing research lacks investigation into the hallucinations of time-series chart summaries generated by LLMs.}

In this paper, we identify important elements for time-series chart summary (e.g., key extremum and upward trend), as well as types of hallucinations in summaries generated by LLMs, such as \textcolor{black}{Trend Direction Error} (i.e., misinterpret an upward trend as a downward trend or the opposite) and Extremum Error (i.e., misjudge the extremum point \textcolor{black}{as a maximum}). To alleviate these \textcolor{black}{hallucinations}, we propose a framework that takes \textcolor{black}{visualization specification} and data table as input, combining external modules, multi-agent iterative collaboration, and self-consistency test to automatically generate summary. This framework integrates natural language reasoning capabilities seamlessly with external tools (data analysis modules), combining the analytical power of language with the computational efficiency of tools to enhance chart summary. We assign multiple agents to engage in generating initial version of chart summary and iterative collaboration, during which they invoke external data analysis modules designed to reduce specific hallucinations, extract data insights, or compile insights into a coherent summary. At last, we use a self-consistency test method to validate and correct our summary, finally arriving at a refined and comprehensive chart summary. We design an interface that assists users in converting charts into summaries using our ChartInsighter, in which we implement an interaction that links text to visualizations, facilitating users to identify potential hallucinations. 

We create a benchmark of 75 pairs of charts and corresponding summaries including a total of 2693 sentences. For a given chart, we generate 4 summaries: one gold summary created manually, one summary generated by ChartInsighter, \textcolor{black}{VL2NL\cite{ko2024natural}, and GPT-4}, with hallucination types annotated at sentence level for all summaries, \textcolor{black}{aiming to evaluate the effectiveness of reducing hallucinations. We develop evaluations based on our benchmark to validate that our method  outperforms state-of-the-art LLMs, uncovers more data insights, produces summaries with richer and more effective semantics, and significantly reduces hallucinations.}

Our main contributions are as follows:
\begin{itemize}
    \item We identify key elements for time-series chart summary and the types of hallucinations produced by LLMs. These serve as guidelines to steer the generation of \textcolor{black}{time-series} chart summary by LLMs.
    \item We propose ChartInsighter to automatically generate time-series chart summaries utilizing iterative fine-grained multi-agent collaboration to arrive at a comprehensive summary. Evaluations show that our system outperforms \textcolor{black}{state-of-the-art} LLMs in generating chart summary and effectively mitigates common hallucinations produced in the generation process.
    \item We create a high-quality benchmark of charts and summaries, with hallucination types annotated, shedding light on further research on reducing hallucinations of summary generation.
\end{itemize}

%% file: sections/2-RelatedWork.tex
\section{Related Work}
Our work builds on prior research on LLMs for visualization, enhancing reasoning and factual knowledge \textcolor{black}{in LLMs, and LLMs for chart summarization}.

%第一节

\subsection{Large Language Models for Visualization}
In recent years, with the rapid development of LLMs, researchers have begun to explore their potential in the field of visualization\cite{yang2024foundation, ye2024generative}. A typical application of LLM4VIS involves generating visual content using a Natural Language Interface (NLI)\cite{narechania2020nl4dv, setlur2019inferencing}. Traditionally, visualization generation relies on machine learning algorithms, utilizing rule-based and constraint-based methods\cite{zhang2023adavis,li2021kg4vis}. With the emergence of LLMs, their powerful text processing capabilities have enhanced tasks such as code generation and storytelling\cite{he2024leveraging}. The LLM4Vis framework\cite{wang2023llm4vis} delivers visualization recommendations based on minimal examples by utilizing feature descriptions, selecting demonstration examples, generating explanations, and outlining reasoning steps to offer human-like interpretations. Similarly, LIDA\cite{dibia2023lida} introduces an innovative tool that automates the generation of visualizations and infographics through a multi-step pipeline, which involves summarization, goal analysis, visualization code generation, and the creation of stylized graphics. While LLM4Vis and LIDA are focused on generating visualizations from datasets, \textcolor{black}{ChartGPT\cite{tian2024chartgpt} and LightVA\cite{lightva} are designed to generate visualizations and extract insights from abstract or ambiguous natural language inputs. LEVA\cite{leva} enables LLMs to understand chart information and the relationships between charts to provide analysis tasks and interaction recommendations.} This indicates that LLMs possess knowledge about visualization and the ability to write visualization code, supporting our work. In contrast to the extensive text-to-visualization research, we focus on the automatic generation of time-series chart summaries.

\begin{figure*}
\includegraphics[width=1\textwidth]{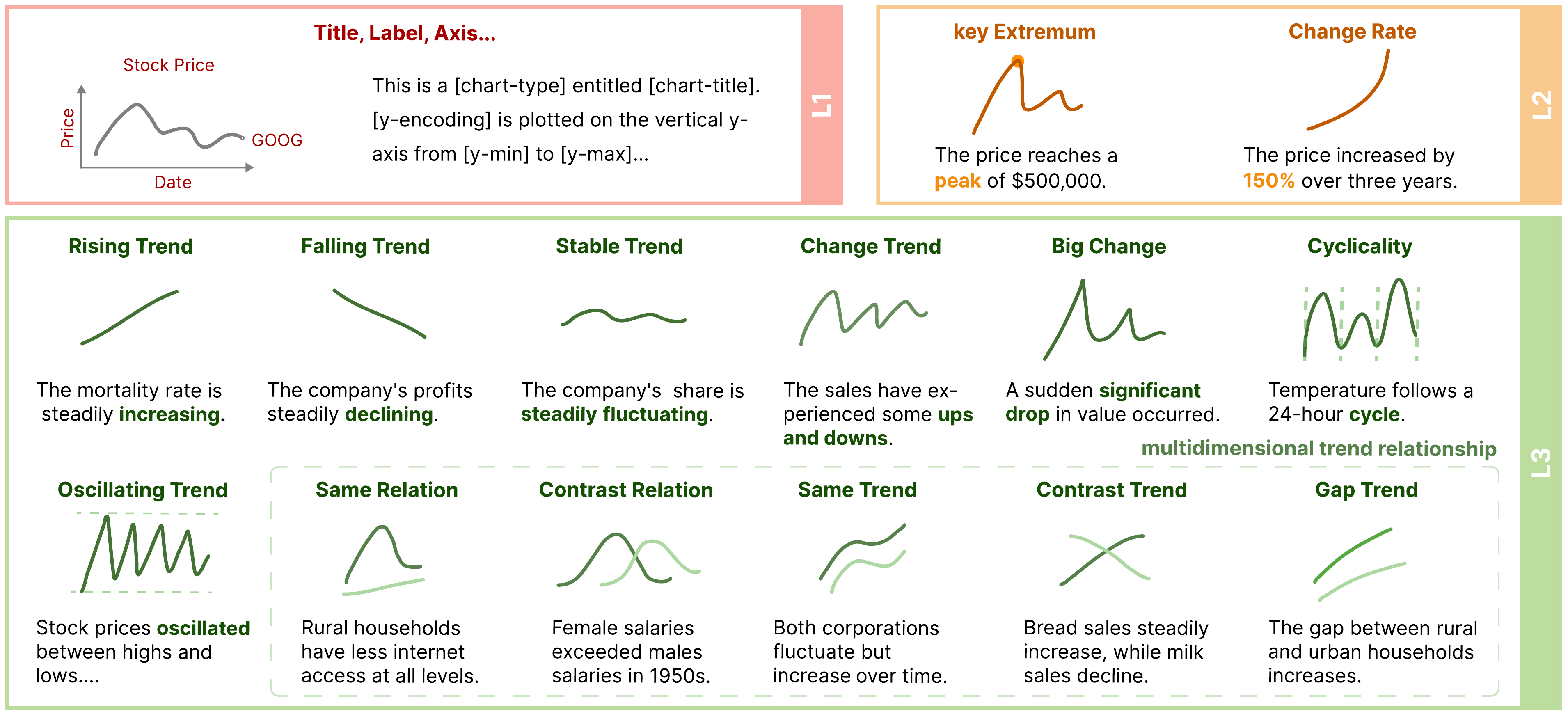}
\caption{\textcolor{black}{Examples of time-series chart summary elements. We classify them into L1-L3, employ simple line diagrams to visually illustrate the meaning of these elements, and present example sentences containing specific elements.}}
\label{fig:Summary Components}      
% Give a unique label
\vspace{-7px}
\end{figure*}

\subsection{Enhancing reasoning and factual knowledge in LLMs}
Researchers raise a number of prompting approaches to enhance LLMs' reasoning ability. Wei et al.\cite{wei2022chain} introduce Chain of Thought (CoT), which guides LLMs to decompose a complex reasoning problem into intermediate steps, and then solve each task step-by-step. \textcolor{black}{Program of Thought (PoT)\cite{chen2023programthoughtspromptingdisentangling} expresses reasoning steps as Python programs, and it leverages the computational capabilities of Python to improve the accuracy of data processing in LLMs.} Although great progress has been made in decreasing hallucinated facts by using CoT, LLMs are still not one hundred percent reliable in handling reasoning tasks\cite{zheng2023does, frieder2024mathematical, yuan2023well}. \textcolor{black}{To further improve the accuracy of the LLM answers,} Wang et al.\cite{wang2022self} propose a self-consistency strategy, \textcolor{black}{which is designed to generate multiple reasoning paths and determine the final answer by employing a majority vote among them.} Self-Contrast\cite{zhang2024self} generates multiple reasoning paths and re-evaluates and revises the text by comparing differences among them. Multi-agent interaction has also been integrated into LLMs to reduce hallucination and augment its problem-solving ability. Cohen et al.\cite{cohen2023lm} asks one LLM to generate a statement and another LLM to check its truthfulness by raising questions. Considering the increased inference cost of leveraging multiple LLMs, Wang et al.\cite{wang2023unleashing} instead propose using a single LLM to simulate and iteratively self-collaborate with different personas. We combine various prompting strategies and employ external modules to ensure analysis accuracy, guiding LLMs to analyze charts step by step according to our proposed guidelines, which greatly reduce hallucinations in time-series chart summary generation.

\subsection{Large Language Models for Chart Summarization}
% \begin{figure*}
% \includegraphics[width=1\textwidth]{figs/Related work.png}
% \summary{Related work}
% \label{framework}      
% % Give a unique label
% \end{figure*}
Lundgard and Satyanarayan\cite{lundgard2021accessible} categorize semantic content into four levels: L1 content includes chart construction(e.g., axis); L2 content describes statistical concepts and relations(e.g., extrema); L3 content refers to perceptual and cognitive phenomena(e.g., data trend); L4 content reveals contextual and domain-specific insights. \textcolor{black}{Our summary generation follows this framework. LLMs possess the potential to generate content across L1 to L4 due to their exceptional text processing capabilities\cite{he2024leveraging, huang2022towards} and excellent ability to organize logic and structure\cite{gao2023chatgpt}.} Consequently, they are widely applied in summary generation. DATATALES\cite{datatales} uses templated prompts to guide LLMs in generating chart summaries. ChartThinker\cite{liu2024chartthinker} improves the logical consistency and accuracy of LLM-generated summaries through CoT prompting and context retrieval, but it still exhibits shortcomings in mathematical analysis. Ko et al.\cite{ko2024natural} generate summaries by guiding LLMs to focus on statistical features and leverage external tools for data analysis, reducing numerical hallucinations in LLMs. However, none of the above chart summaries include L3 content. VisText\cite{tang2023vistext}, through fine-tuning, enables LLMs to generate summaries including L1-L3 content, \textcolor{black}{but it only focuses on handling simple, uni-dimensional
charts and cannot extract relationships between data from different dimensions in multidimensional data}. Therefore, there is a lack of work that simultaneously addresses the computation and reasoning limitations of LLMs to guarantee both correctness and semantic richness in the automation of summary generation. To bridge this gap, we propose a new framework for generating time-series chart summaries.

%% file: sections/3-Preliminaries.tex
\section{Preliminaries}
\label{sec:Preliminaries}
In this section, we derive the requirements for generating an accurate and comprehensive summary of the time-series data chart. We summarize the key summary elements according to L1-L3 content categorization\cite{lundgard2021accessible}. We also test real-world data using \textcolor{black}{state-of-the-art} LLMs to identify the types of hallucinations that LLMs may produce during summary generation.

% 3.1
\subsection{Requirements}
Based on prior literature, we identify two key requirements for generating accurate and effective time-series chart summary generation.

\textbf{Analyze structural summary elements of time-series charts.} 
While many previous works have focused on automatically generating complete summaries based on large models\cite{masry2023unichart,tang2023vistext, xia2024chartx, kantharaj2022chart,obeid2020chart}, they often lack a fine-grained understanding of content in L1-L3 for time-series data. To address this, we need to conduct a detailed analysis and refine the elements for L1-L3, ensuring a more precise and granular understanding of the content at these levels, and thus a more comprehensive and structured summary.

\textbf{Summarize hallucination types of time-series chart summary generation.} 
Previous research has shown that various hallucinations can occur when automatically generating chart summaries\cite{tang2023vistext, obeid2020chart, obaid2023tackling}. By categorizing these different types of hallucinations, we can more clearly identify and understand the characteristics of each type, allowing us to develop more targeted solutions. Since different types of hallucinations may require different approaches, classification helps us accurately pinpoint the issues and make quick adjustments to improve the accuracy and reliability of the generated summaries. Thus, we need to summarize the hallucination types when generating a chart summary.\par

% 3.2

\subsection{Summary Elements}
\label{sec:Summary Elements}
To generate comprehensive time-series chart summaries, we have summarized the essential elements from existing research\cite{kim2021towards,greenbacker2011abstractive,kim2018multimodal,burns2009modeling,kim2023emphasischecker,shi2023supporting,law2020characterizing,ma2021metainsight} and real-world chart dataset websites\cite{pewresearch, ourworldindata} and categorized them into L1 to L3 (Fig.~\ref{fig:Summary Components}).\par

\textbf{L1} content includes elemental and encoded properties\cite{lundgard2021accessible}, such as \textit{Title}, \textit{Label} and \textit{Axis}, which describe the visual elements of the chart's construction.
In \textbf{L2} content, \textcolor{black}{the most common key terms in the summaries of time-series charts are \textit{Key Extremum} and \textit{Growth Rate}. The \textit{Key Extremum} is the key turning points of the curve, indicating significant events.} \textcolor{black}{The \textit{Growth Rate} quantitatively describes the speed or intensity of changes in time-series data.}

\textcolor{black}{The content in \textbf{L3} is divided into two main categories: unidimensional trend description and multidimensional trend relationship description. For unidimensional trend description, the main elements are}: \textit{Rising Trend} and \textit{Falling Trend}, the most common; \textit{Stable Trend}, indicating little change; \textit{Change Trend}, which fluctuates up and down; \textit{Big Change}, referring to sharp increases or decreases; \textit{Cyclicality}, describing repeating patterns over time; and \textit{Oscillating Trend}, which fluctuates between multiple levels. \textcolor{black}{For multidimensional trend relationship description,} \textit{Same Relation} means entities maintain a consistent relationship (e.g., line 1 is always above line 2); \textit{Contrast Relation} indicates a shift in dominance (e.g., line 1 leads in the first half, but line 2 overtakes in the second half); \textit{Same Trend} means both dimensions move in the same direction (either both rising or both falling); \textit{Contrast Trend} refers to one dimension rising while the other falls; and \textit{Gap Trend} indicates the gap between entities changes consistently in one direction.

\subsection{Hallucination Types}
\label{sec:Hallucination Types}

To gain a comprehensive understanding of the potential hallucinations that may occur when LLMs generate time-series chart summaries, \textcolor{black}{we conducted tests using four state-of-the-art LLMs: GPT-4\cite{achiam2023gpt}, Claude-3\cite{claude2023}, GPT-4o\cite{openai_gpt4_2024}, and LLaMA-3.1-70B\cite{llama}.} We sourced 20 \textcolor{black}{time-series line} charts from real-world datasets\cite{ourworldindata,pewresearch, stlouisfed}. These charts included both unidimensional and multidimensional time-series data covering diverse fields such as finance, energy, politics, education, and environment. \textcolor{black}{We prompted LLMs to generate L1-L3 summaries, using Vega-Lite specification\cite{satyanarayan2016vega} and data table as the input.} Each chart was summarized by all the aforementioned models, resulting in a total of 80 summaries, containing 1083 sentences in total, \textcolor{black}{and among them, 199 instances of hallucinations were identified. Then four authors, all with visualization backgrounds, reviewed the L1-L3 parts of the generated summary and independently created initial classifications of hallucination types. Then, we integrated each person's classifications and collectively discussed the different findings to collaboratively establish a unified, final taxonomy. During this process, we re-examined the summaries to ensure that all hallucination types were accurately classified. If there were any discrepancies in our classifications, we engaged in in-depth discussions until a consensus was reached, ensuring the accuracy of the classification results.}

\begin{figure}[htbp]
    \centering
    \includegraphics[width=1\linewidth]{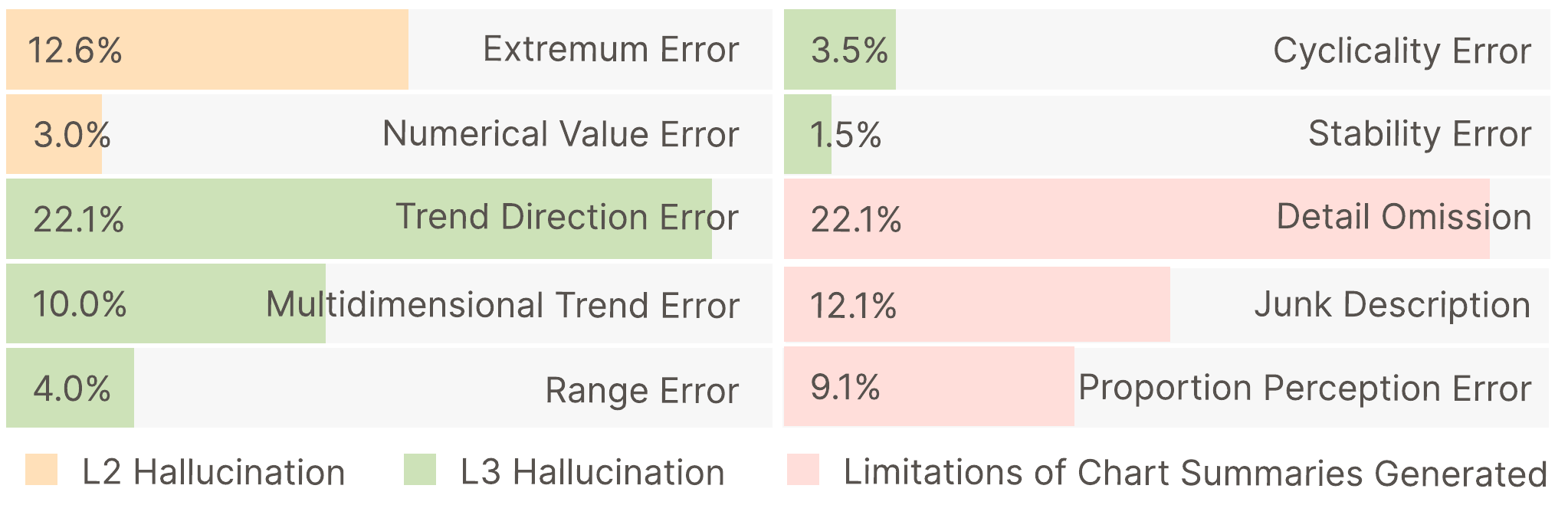}
    \caption{The frequency of different types of hallucinations in LLM-generated time-series chart summaries.}
    \label{hallucination}
    \vspace{-5px}
\end{figure}

\textcolor{black}{We have identified a total of 10 types of hallucinations in Fig.~\ref{hallucination}. We found that none of the hallucinations occurred in the L1 summary element generation based on our test. Therefore, we have classified them as L2 and L3, and summarized the limitations of LLM-generated time-series chart summaries.}

For \textbf{L2} hallucinations, we have concluded the following 2 types:

\textbf{Extremum Error.} 12.6\% of 199 instances of hallucination erroneous statements include it. This error occurs when LLMs incorrectly describe a local extremum as the absolute maximum or minimum, when in fact it is just a regular peak or trough value\textcolor{black}{, or mistakenly identify an ordinary value as an extremum.}

\textcolor{black}{\textbf{Numerical Value Error}. 3.0\% of statements include such error. This error occurs when there is a discrepancy in describing or calculating quantitative data. The rarity of this error stems not from the fact that the LLM has strong numerical computation capabilities, but from the fact that it rarely includes insights that require numerical calculations in its summaries, thus not exposing this issue much.}

For \textbf{L3} hallucinations, we have categorized them into 5 types:

\textbf{Trend Direction Error.} 22.1\% of the erroneous statements include this error. This error arises when LLMs incorrectly identify the direction of a trend, such as misinterpreting an upward trend as a downward one, or vice versa.

\textbf{Multidimensional Trend Error.} 10.0\% of error instances occur, where trends across multiple dimensions are misinterpreted. \textcolor{black}{LLMs either mistake the same trends/relations as contrast, or conversely, mistake contrast trends/relations as the same. When describing multidimensional trends, they mix data insights from different dimensions together, leading to a very confusing and disorganized presentation. For example, LLMs combine two dimensions into one, like ``Google’s stock price rose before 2010 and peaked in 2012''. However, ``peaked in 2012'' is the attribute of another dimension, not Google.}

\textcolor{black}{\textbf{Range Error.} There are 4.0\% cases of this error. When analyzing time-series data, the start and end times of trends are incorrectly identified. When a sudden trend reversal occurs, LLMs fail to promptly recognize and adjust to changes in the data, leading to an incorrect description of the trend. }

\textbf{Cyclicality Error.} There are 3.5\% instances of this type of error, where non-cyclical trends were incorrectly interpreted as cyclical.

\textcolor{black}{\textbf{Stability Error.} This error, accounting for 1.5\% of total errors, occurs when a fluctuating trend is incorrectly described as stable, or when stable data is misrepresented as fluctuating, leading to a skewed perception of the actual trend.}

We have categorized the following 3 classes as \textbf{Limitations of Chart Summaries Generated}, which are common problems in L2 and L3:

\textbf{Detail Omission.} We identified 22.1\% instances of this error, making it the most common. \textcolor{black}{This error refers to when LLMs tend to generalize data within a specific range, focusing on overall trends while overlooking key fluctuations and turning points in time-series data. This oversight results in the masking of crucial underlying information, leading to an incomplete understanding of the data and potentially affecting the final reasoning and decision-making. Additionally, when describing multidimensional time-series line charts, LLMs tend to focus on data from a single dimension, overlooking the others in summary.}

\textbf{Junk Description.} There are 12.1\% examples including such a problem. 
This drawback can take the form of broad generalizations that fail to specify key details, such as saying ``some countries grow faster and others slower'' without naming the countries, or frequent mention of various numerical values that represent meaningless points. It can confuse the reader and reduce the effectiveness of the description.

\textbf{Proportion Perception Error.} This error accounts for 9.1\%. When describing fluctuations, \textcolor{black}{terms like} ``significant'' are often inaccurately used, even if the magnitude of these fluctuations is quite minor compared to other parts of the same line or to fluctuations in other lines. This error highlights a common issue where LLMs fail to appropriately scale its descriptions relative to the overall data variability.

\begin{figure*}[t]
\includegraphics[width=1\textwidth]{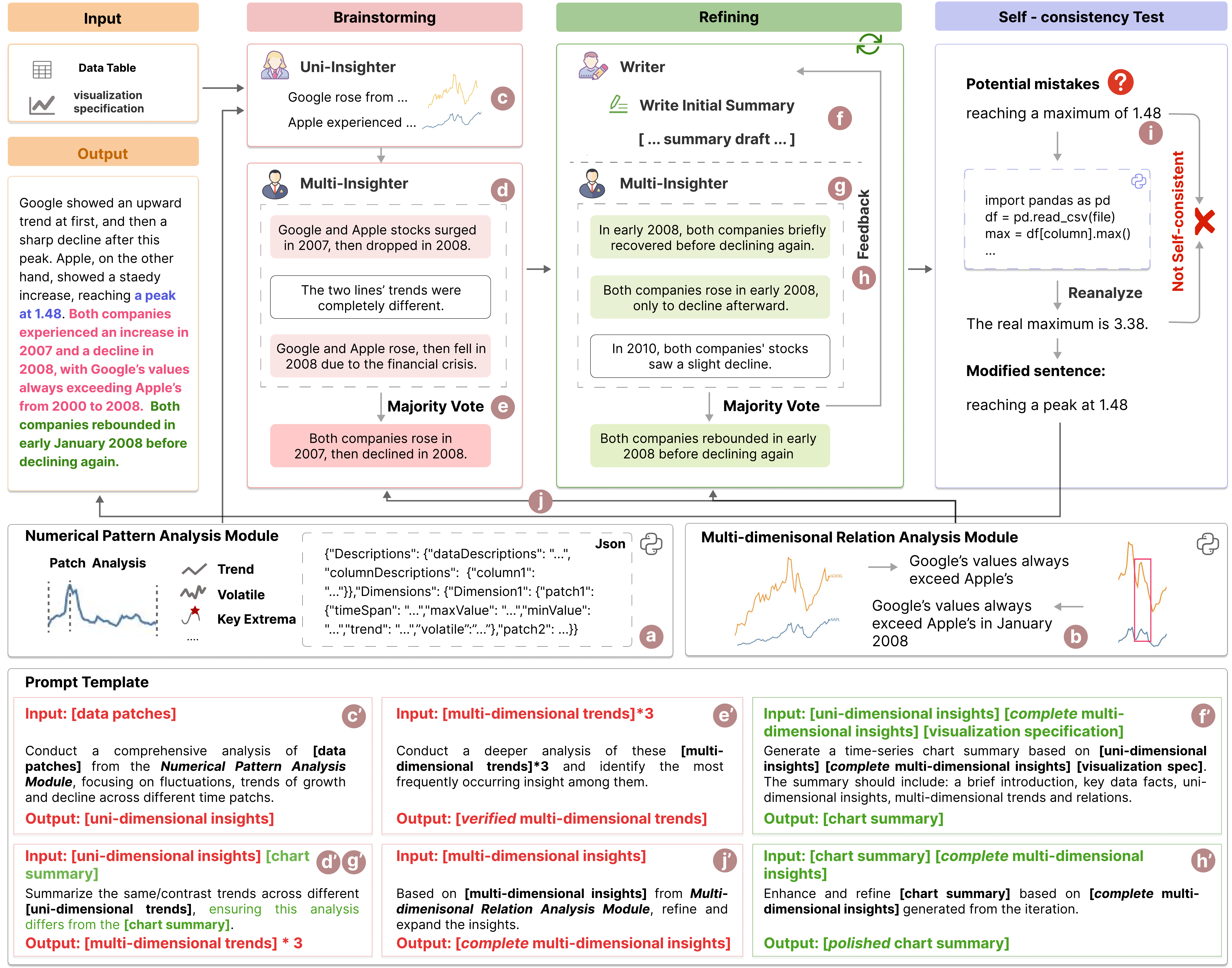}
\caption{\textcolor{black}{The pipeline of ChartInsighter includes three steps: Brainstorming, Refining, and Self-consistency Test.} In ChartInsighter, we input visualization specification and data table to initiate the analysis process. This is first handled by both Uni-Insighter and Multi-Insighter which generate preliminary uni- and multi-dimensional data insights respectively, and compile an initial summary. In the refining stage, we have designed a multi-agent collaborative process between the Multi-Insighter and the Writer. This iterative process, which involves both mining and organizing insights, enables us to achieve a relatively accurate and comprehensive summary. At last, in the self-consistency test phase, we concentrate on identifying and addressing key types of hallucinations to produce the final version of chart summary. In Prompt Template, we display the input, prompt, and output of each step. For example, the input, prompt, and output of Step \textit{c} are demonstrated in Prompt Template \textit{c'}. It should be specifically pointed out that Step \textit{g} builds upon the input and prompt from Step \textit{d}, with additional new content highlighted in orange font in Prompt Template \textit{g'}.}
\label{fig:Pipeline}      
% Give a unique label
\vspace{-8px}
\end{figure*}

% \begin{figure*}[t]
% \includegraphics[width=1\textwidth]{figs/Prompt.pdf}
% \caption{\textcolor{red}{The figure shows the prompts received by Uni-Insighter, Multi-Insighterr and Writer, as well as the inputs and outputs at each step in the Brainstorming and Refining.}}
% \label{fig:prompt}      
% % Give a unique label
% \end{figure*}

Statistics indicate that among all types of hallucinations, the most common ones are \textit{Detail Omission}, for which it is challenging for LLMs to generate a comprehensive summary that covers every dimension and all key points, particularly for a complex chart, and \textit{Trend Direction Error}. Other frequent errors include \textit{Extremum Error}, and \textit{Range Error}, stemming from poor semantic understanding of context. To address these, we use mathematical calculations in the \textcolor{black}{external module} to improve data analysis accuracy. \textit{Multidimensional Trend Error} is another common issue, which we mitigate by \textcolor{black}{employing majority voting to select the most frequent multidimensional insight.} Additionally, we find that when LLMs call external tools to generate code to analyze data, they tend to focus only on basic metrics like averages, extremes, and growth rates, often failing to provide a complete, logical chart summary, while our framework effectively mitigates this problem.

These hallucinations severely impact the reader's understanding of the charts, causing significant confusion. Therefore, our framework mitigates these hallucinations to ensure clearer and more accurate summaries, especially \textit{Extremum Error}, \textit{Numerical Value Error}, \textit{Trend Direction Error}, \textit{Multidimensional Trend Error}, \textit{Range Error}, \textit{Detail Omission}, \textit{Junk Description} and \textit{Proportion Perception Error}.

%% file: sections/4-ChartInsighter.tex
\section{ChartInsighter}

\textcolor{black}{
Based on the guidelines in Sec.~\ref{sec:Preliminaries}, we propose a pipeline for generating time-series chart summaries and an interface to support summary generation, in which we implement an interaction that links text to the chart. In our pipeline, we employ a multi-agent iterative collaboration, along with external modules for data analysis.}

\subsection{Brainstorming}
In this step, we assign two agents, \textit{Uni-Insighter} and \textit{Multi-Insighter}. After inputting the time-series data and visualization specification, \textit{Uni-Insighter} analyzes and generates uni-dimensional insights for each dimension (e.g., extrema, trends). And based on these uni-dimensional descriptions output by \textit{Uni-Insighter}, \textit{Multi-Insighter} generates multi-dimensional insights.

In \textit{Uni-Insighter}, we have designed \textbf{Numerical Pattern Analysis Module} to divide long time-series data into smaller data patches. \textcolor{black}{We determine the segmentation points of the time-series by identifying significant extreme values in the time-series data, thereby dividing the data into multiple patches, each with relatively consistent trend changes. To further optimize the segmentation, we merge consecutive patches that exhibit minimal fluctuation changes. We characterize the volatility of each patch using its variance and establish a threshold based on the median of the variances across all patches,  adding $k$ times the standard deviation of these variances\cite{HowTallIsTall}. After multiple attempts, we found that setting the k-value to 0 yielded the best results. Subsequently, multiple consecutive patches whose variance falls below this threshold are merged to finalize the segmentation. This process ensures that the number of patches is minimized while maintaining consistent trend patterns within each patch. We then calculate key statistics (e.g., max, min, volatility) for different patches, along with providing a semantic description of the data (Fig.~\ref{fig:Pipeline}-a). These statistic features are designed to create an information-dense and compact representation of the data, which helps LLMs better understand the variations in time-series data\cite{dibia2023lida}.} This module specifically uncovers uni-dimensional insight discussed in Sec.~\ref{sec:Summary Elements}.

In \textit{Multi-Insighter}, we have designed \textbf{Majority Vote} and \textbf{Multi-dimensional Relation Analysis Module}. \textit{Multi-Insighter} generates multi-dimensional trend description based on the output of \textit{Uni-Insighter}, repeating the process \textcolor{black}{multiple times to elicit
diverse descriptions. We empirically set the repeating times to three to ensure a balanced outcome. It then applies Majority Vote to filter and validate the generated trend description, which leverages \textit{Multi-Insighter} to select the most consistent answer among multiple candidates\cite{chen2023universal}.} Multi-dimensional Relation Analysis Module analyzes the insights obtained through Majority Vote and outputs insights, which have supplemented the multi-dimensional relation and refined the temporal precision. Then they are delivered to \textit{Multi-Insighter} back to generate complete multi-dimensional insights. The implementation details of Multi-dimensional Relation Analysis Module are as follows: we begin by determining whether intersections exist between multiple dimensions, enabling us to identify whether they are of the same relation or contrast relation. If the same relations are identified, we proceed to evaluate the rankings of the dimensions across various time periods. Based on the results of the previous analysis, we can pinpoint the corresponding time periods. To be specific, we first utilize LLM to identify temporal expressions within the sentence, such as specific time points (e.g., ``mid-20th century'') or durations (e.g., ``in two years''). Then LLM will determine the relevant period according to the context and analysis result. Next, we match these temporal expressions to the corresponding data. This process presents a challenge because the temporal descriptions in the sentence are often coarse (a sentence may specify a time range such as ``2000-2024'', while the actual data is recorded at a finer granularity, such as daily or monthly intervals). Consequently, directly mapping these temporal descriptions to the raw long time-series data proves difficult. However, the broader time ranges mentioned in the sentence are based on the patches divided by Numerical Pattern Analysis Module, so we can directly pinpoint a more precise time point according to the patches. The output result is shown in Fig.~\ref{fig:Pipeline}-b.

\begin{figure*}
\includegraphics[width=1\textwidth]{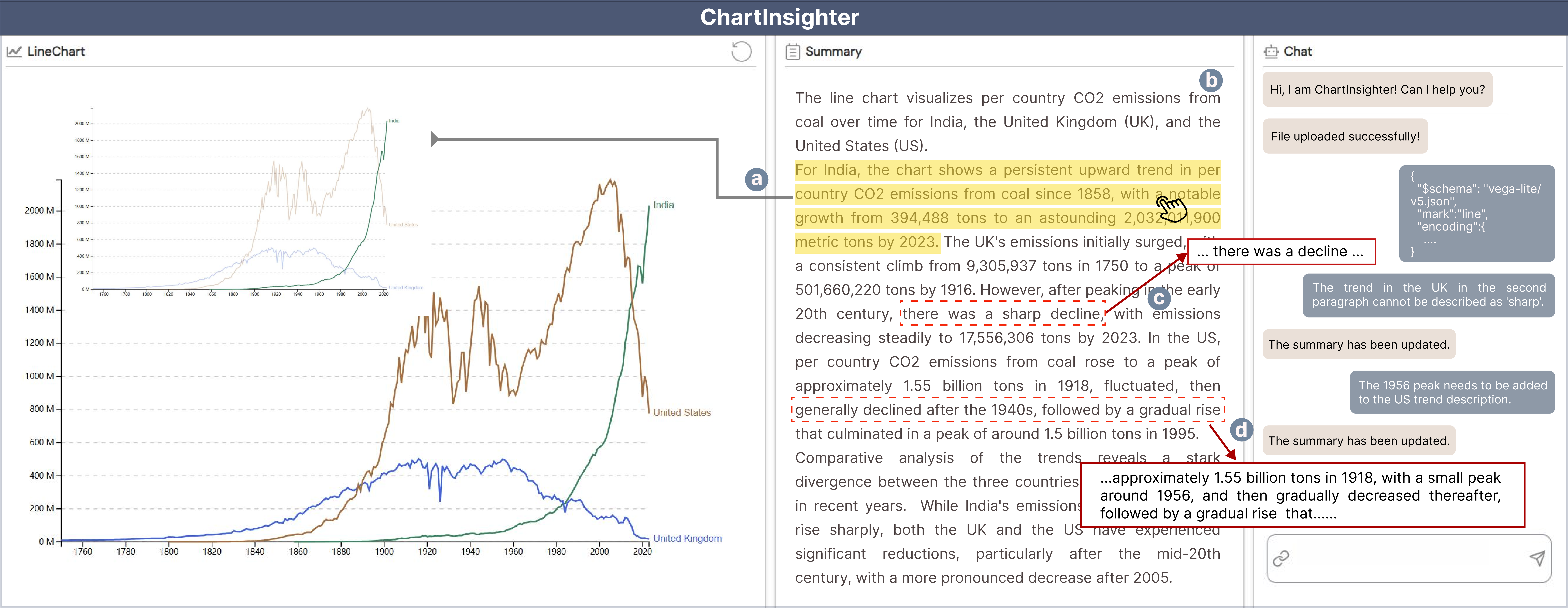}
\caption{\textcolor{black}{The overview of ChartInsighter. Users can input a Vega-Lite specification and data table to generate a summary. By hovering over sentences containing data references, the corresponding portions in the chart are highlighted (a). Additionally, users can interact with the chat view, prompting the model to modify the summary or elaborate on details they find more interesting, resulting in a more satisfactory summary.}}
\label{fig:Interface}      
% Give a unique label
\vspace{-8px}
\end{figure*}

In Fig.~\ref{fig:Pipeline}, we demonstrate our workflow using a real case. After inputting Google's and Apple's \textcolor{black}{stock data from 2000 to 2010,} along with the corresponding visualization \textcolor{black}{specification}, \textit{Uni-Insighter} first invokes Numerical Pattern Analysis Module to analyze the data, generating a JSON (Fig.~\ref{fig:Pipeline}-a). Based on the JSON, \textit{Uni-Insighter} creates individual, one-dimensional trends for the two companies (Fig.~\ref{fig:Pipeline}-c) with the prompt in Fig.~\ref{fig:Pipeline}-c'. Subsequently, \textit{Multi-Insighter} examines the uni-dimensional trends to determine whether the two dimensions have the same or contrast trends (output in Fig.~\ref{fig:Pipeline}-d, prompt in Fig.~\ref{fig:Pipeline}-d'). This analysis is repeated three times, thus generating three multi-dimensional data insights, with majority voting (Fig.~\ref{fig:Pipeline}-e') confirming the result (Fig.~\ref{fig:Pipeline}-e). The analysis concludes that both companies exhibit the same trend: an increase in 2007 followed by a decline in 2008. This insight is forwarded to Multi-dimensional Relation Analysis Module, which performs further analysis and concludes that Google’s values consistently exceed Apple’s during this period(Fig.~\ref{fig:Pipeline}-b). Finally, \textit{Multi-Insighter} synthesizes all insights with the prompt in Fig.~\ref{fig:Pipeline}-j' and concludes that both companies experienced an increase in 2007 and a decline in 2008, with Google’s values always exceeding Apple’s from 2000 to 2008.

\subsection{Refining}
In this step, \textit{Multi-Insighter} and \textit{Writer} collaborate through multiple rounds to produce a smoother and more comprehensive summary. \textit{Writer} first generates an initial summary based on the uni-dimensional insights from \textit{Uni-Insighter}, the multi-dimensional insights from \textit{Multi-Insighter}, and the visualization specification. \textit{Multi-Insighter} then supplements and refines the initial summary using the same approach as the one in Brainstorming. \textit{Writer} continues to integrate the additional content provided by \textit{Multi-Insighter}, and \textit{Multi-Insighter} supplement again, iterating through multiple rounds.

We use Fig.~\ref{fig:Pipeline} to demonstrate the iteration process. After Brainstorming, \textit{Multi-Insighter}’s multi-dimensional insights, \textit{Uni-Insighter}’s one-dimensional trends \textcolor{black}{and visualization specification} are sent to the \textit{Writer} to draft an initial summary (Fig.~\ref{fig:Pipeline}-f) with the prompt in Fig.~\ref{fig:Pipeline}-f'. After completing the draft, \textit{Writer} returns it to \textit{Multi-Insighter} for further review. \textit{Multi-Insighter} reanalyzes the insights and uses majority voting to confirm additional findings, such as the observation that both companies rebounded in early 2008 before declining again (Fig.~\ref{fig:Pipeline}-g). The system calls Multi-dimensional Relation Analysis Module again, which pinpoints the timeline to January 2008. \textit{Multi-Insighter} synthesizes these insights (Fig.~\ref{fig:Pipeline}-h') and provides the updated summary to the \textit{Writer} for further refinement. This iterative process continues until \textit{Multi-Insighter} determines that no new insights remain uncovered. The final output is a comprehensive summary.

\subsection{Self-consistency Test} We then conduct a Self-consistency test to detect potential errors in the generated chart summary and output a corrected one. In the previous steps, we have called external modules to mitigate some hallucinations (e.g., Trend Direction Error and Multi-dimensional Trend Error). Since our summary is derived through patch-based analysis, LLMs may mistakenly identify local extrema within individual patches as global extrema. This causes the LLM to incorrectly describe the identified data insights using terms like ``maximum'' which are actually not. Similarly, fluctuations that appear significant within a specific patch may lose prominence when evaluated across the entire curve. Therefore, it would be inappropriate to characterize these fluctuations as significant within a broader context.

To address these issues, we focus specifically on detecting and correcting Extremum Error and Proportion Perception Error in the summary. We guide the LLM to identify these potential errors and reanalyze the questionable sentences. Then we  \textcolor{black}{instruct} LLM to compare the newly generated sentences with the original sentences. If the sentences are consistent, it confirms that our original summary is accurate, and we output the original summary as the final version. However, if the data insights conveyed in sentences differ, LLM \textcolor{black}{revises 
 the original sentences.}

In Fig.~\ref{fig:Pipeline}, we can see that this step identifies a potential error in the sentence ``Apple reached a maximum of 1.48'' (Fig.~\ref{fig:Pipeline}-i). Then LLM checks and figures out that the maximum should be 3.38. Then the sentence is revised accordingly \textcolor{black}{and the final summary is output}.

\subsection{Linking Summary to Chart}
\textcolor{black}{We design a system to help users effectively generate time-series chart summaries and allow them to personalize and modify the generated summaries. Our interface consists of 3 parts: chart view, summary view, and chat view (Fig.~\ref{fig:Interface}). The interaction process begins with the chat view, where users provide the data and visualization specification.} The chart view then generates the corresponding chart visualization, and a preliminary version of the summary is automatically created \textcolor{black}{in summary view. Users can further refine the summary based on their needs until finally arriving at a correct and satisfactory version.}\par

Inspired by \cite{datatales,kim2023emphasischecker}, we link the text in the generated summary to the chart in interaction. Sentences containing data \textcolor{black}{references} are underlined with a dashed line, and when users hover over these sentences, the corresponding portion in the chart is highlighted (Fig.~\ref{fig:Interface}-a).

\begin{figure*}[t]
\begin{center} % 仅让图片居中
\includegraphics[width=1\textwidth]{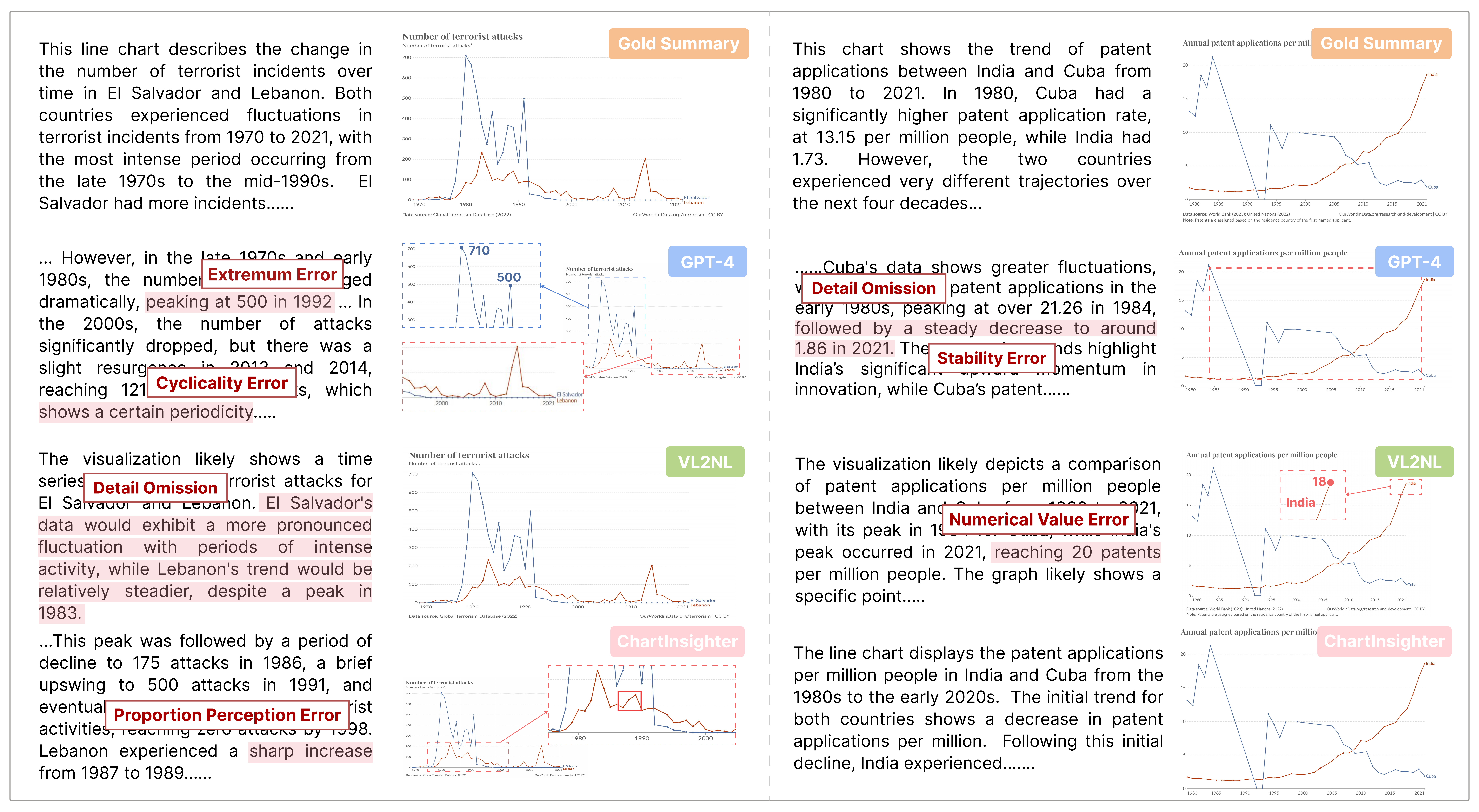}
\end{center}
\caption{\textcolor{black}{An example of our benchmark dataset.} We carefully crafted gold summaries and labeled the hallucinations at sentence granularity for the summaries generated by each of the three models GPT-4, VL2NL, and ChartInsighter for each chart.}
\label{fig:bm}      
% Give a unique label
\vspace{-8px}
\end{figure*}

This interactive linking design helps users quickly map the text to the relevant portions in the chart, especially in complex, multi-dimensional charts, significantly reducing the time and effort needed to locate specific chart portion. Additionally, it facilitates checking for any potential hallucinations in the summary.

%% file: sections/5-Benchmark.tex
\section{Benchmark}

LLMs often generate chart summaries that contain significant hallucinations \cite{tang2023vistext}, making the mitigation of hallucinations an important task. In the future, there will be numerous research efforts aimed at reducing these hallucinations.

To bridge this gap, we have introduced a benchmark for time-series chart summary generation. First, we have systematically summarized a set of hallucination types and their definitions that occur when LLMs generate summaries for time-series data, \textcolor{black}{as detailed in sec.~\ref{sec:Hallucination Types}}. This framework enables a clearer and more quantifiable understanding of the shortcomings in LLMs' ability to produce accurate and reliable summaries. \textcolor{black}{Second, we have constructed a benchmark that, for each chart (a total of 75 charts), includes data, Vega-Lite specification, chart image, a gold summary created manually, and three summaries generated by GPT-4, VL2NL, and ChartInsighter. We have annotated the types of hallucinations present in each sentence (Fig.~\ref{fig:bm}), which was then utilized to perform a comparative analysis of the generation performance between these three models. We have also evaluated and quantified key metrics, including hallucination rate and semantic richness, which were subsequently utilized to perform a comparative analysis of the generation performance among
these three models, as detailed in sec.~\ref{qualityEvaluation}. Future research can use our benchmark to evaluate the effectiveness of their hallucination mitigation techniques.}

%\textcolor{black}{Our benchmark dataset possesses the following qualities. 1) \textit{High-quality data}: the dataset's summaries, visualization specifications, and other components are carefully created and validated through multiple rounds to ensure accurate representation of key time-series data information. 2) \textit{Accurate annotation of hallucinations}: our dataset includes precise annotations of hallucinations, and each hallucination has been rigorously reviewed to ensure reliability and objectivity. 3) \textit{Wide coverage}: the dataset spans a broad range of domains, including charts of varying complexity. The dataset construction process is shown below:}

\textcolor{black}{We collected 75 time-series line charts from reputable real-world datasets\cite{ourworldindata,stlouisfed,vegalite,statista}, covering domains such as economics, environment, and energy, with 25 charts for each of three complexity levels: simple, moderate, and complex. These levels were determined through consensus after extensive discussion among three of the authors, ensuring the comprehensiveness of the dataset.}
We assess chart complexity based on several data features, including peak values, sequence length, dimensions, and variation patterns. First, we consider the number of peaks and the distance between peaks and valleys. Data with large differences between peaks and valleys, and frequent fluctuations, are more complex. Simple charts typically have 1-2 significant peaks, moderate ones have 3-4, and complex ones have \textcolor{black}{more than 4 peaks.} Next, the sequence length and data dimensions are evaluated. Longer sequences and higher dimensions increase chart complexity. \textcolor{black}{However, even if a chart has five peaks but only one dimension, we still categorize it as a complex chart due to the high number of fluctuations.} Lastly, we examine the variation pattern. If the data shows periodicity or trends, the chart is less complex. Irregular or unpredictable data increases complexity.

For each chart, we recruited 6 participants aged 23-28 with a background in data visualization to create gold summaries covering L1-L3 content. Participants were provided with the chart, data, and guidelines specifying L1-L4 content, and were instructed to focus on describing L1-L3 content. Two approaches were used: if an expert-level summary \textcolor{black}{of the chart} existed, participants refined it based on the guidelines; if not, they used LLMs to generate an initial draft, which they then edited. While human-written summaries typically highlight visually prominent features like peaks and may omit some details, they generally cover L1-L3 content thoroughly.

We used the Vega-Lite specification and the data table of each chart as inputs to guide our model, VL2NL\cite{ko2024natural}, and GPT-4 to generate chart summaries.  \textcolor{black}{We explained the types of hallucinations and their definitions to 6 participants, who then performed a sentence-by-sentence review of each generated summary, annotating instances of hallucinations. Participants were compensated \$10 per hour. We further calculated the frequencies of hallucinations in each summary. To ensure the quality of our benchmark, we conducted a manual review, including verifying the accuracy and completeness of the summaries, validating the Vega-Lite specifications, and examining the classification of hallucination types.} We strive to ensure that the benchmark meets high standards in all dimensions, thereby providing researchers with a high-quality and reliable dataset that supports future applications.

% In our benchmark compressed package, we categorize charts by complexity into three levels: simple, moderate, and complex. In each complexity level's folder, we place pictures of the chart, Vega-Lite code, and data in CSV files. Additionally, we provide a gold summary and the summaries generated by ChartInsighter, GPT-4, and VL2NL with input as Vega-Lite code and data, which are placed in their respective folders. Different folders use different numerical identifiers to mark that these files belong to which chart.
% Moreover, we have a file named simple/middle/complex.xlsx, which records the hallucination of summaries generated by the three models for charts of corresponding complexities. Within each model, we sequentially display the complete summary text, sentence number, L2, L3 number sentences, number of hallucinations,  hallucinatory sentences and their types, human rating, semantic richness, and hallucination rate of each chart.

%% file: sections/6-Evaluation.tex
\section{Evaluation}

\textcolor{black}{In this section, we evaluate the diversity, accuracy, and hallucination rate of the generated summaries, and assess the algorithm's performance, all based on our benchmark.  Finally, a usage scenario is presented to illustrate how ChartInsighter can help the user generate a satisfactory summary.}

\begin{table*}[t]
\centering
\fontsize{8pt}{10pt}\selectfont
\renewcommand{\arraystretch}{1.2}
\setlength{\tabcolsep}{4pt}  % 控制列间距
\begin{tabular}{lccccccc|cc}  % 保留列定义中的竖线
\toprule
\multicolumn{1}{c}{} & 
\multicolumn{7}{c}{\textbf{Automatic \& Human Evaluation}} & 
\multicolumn{2}{c}{\textbf{Quality Evaluation}} \\ \midrule
\textbf{Summary} & \textbf{RC} $\uparrow$ & \textbf{Chamfer} $\uparrow$ & \textbf{MST} $\uparrow$ & \textbf{Span} $\uparrow$ & \textbf{Sparness} $\uparrow$ & \textbf{Entropy} $\uparrow$ & \multicolumn{1}{c}{\textbf{Human} $\uparrow$} & \multicolumn{1}{c}{\textbf{Semantic Richness} $\uparrow$} & \textbf{Hallucination Rate} $\downarrow$ \\ 
\midrule
GOLD  & 1.34  & 1.08  & 17.21 & 0.95 & 1.09 & 2.68 & - & - & - \\
VL2NL & 1.34  & 1.18  & 10.85 & 0.93 & 1.10 & 2.26 & 1.70 & 0.33 & 1.63 \\
GPT-4 & \textbf{1.36} & \textbf{1.21}  & 15.21 & 0.95 & 1.16 & 2.51 & 2.86 & 0.74 & 0.48 \\
OURS  & \textbf{1.36}  & 1.17  & \textbf{25.15} & \textbf{0.97} & \textbf{1.18} & \textbf{3.01} & \textbf{3.79} & \textbf{0.75} & \textbf{0.14} \\
\bottomrule
\end{tabular}
\caption{\textcolor{black}{Evaluation results for different models using our benchmark. We compare VL2NL\cite{ko2024natural} and GPT-4 (our base model) across multiple metrics, including Automatic \& Human Evaluation and Quality Evaluation. $\uparrow$: Higher is better, $\downarrow$ : Lower is better. \textbf{Bold}
represents the best result.}}
\label{table_combined}
\vspace{-8px}
\end{table*}

\subsection{Automatic \& Human Evaluation}
To evaluate our model’s effectiveness in generating chart summaries, we compared it against a gold summary, GPT-4 (our base model), and VL2NL (which also uses Vega-Lite specification input and generates L1-L2 content summaries) \textcolor{black}{using our benchmark to} assess text diversity. We used six evaluation metrics from previous studies\cite{ko2024natural} for automatic evaluation: \textcolor{black}{remote-clique (average of mean pairwise distances), Chamfer distance (average of minimum pairwise distances), MST dispersion (sum of edge weights of MST), span (Pth percentile distance to centroid), sparseness (mean distance to medoid), and entropy (Shannon-Wiener index for points in a grid partition). To assess ChartInsighter’s performance, we calculated the average score for each metric.}

The evaluation results, as shown in Tab.~\ref{table_combined} (Automatic \& Human Evaluation) indicate that ChartInsighter generally generates semantically richer summaries compared to VL2NL and GPT-4. We score the highest in RC, MST, Span, Sparseness, and Entropy, respectively  1.36, 25.15, 0.97, 1.18, and 3.01, and our Chamfer score is 1.17, just slightly below the GPT-4's score of 1.21, both suggesting that our summaries are more dispersed, varied, and complex. However, these metrics may be incomplete and may not fully capture the quality of the summaries, as the scores for the gold summaries in Tab.~\ref{table_combined} are not the highest. Therefore, it is necessary to conduct a human evaluation.

To evaluate the quality of summaries generated by ChartInsighter, we conducted a human evaluation. Six participants, aged 20 to 25, who had experience in reading and writing chart summaries, took part. \textcolor{black}{Before the experiment, we clarified the definitions of accuracy, coverage, summary elements, and types of hallucinations (as outlined in Sec.~\ref{sec:Preliminaries}) to ensure objective evaluation. Participants evaluated each summary based on three criteria: Accuracy, Fluency, and Matching Degree (the matching degree between the chart and the summary).
Each summary was rated on a scale from 1 to 5, with 1 being the lowest and 5 the highest.} Summaries were presented randomly, and the final rating was the average of all ratings. Tab.~\ref{table_combined} shows the average ratings for the 3 summary groups. ChartInsighter received the highest rating of 3.79, indicating it meets user needs most effectively.

\subsection{Quality Evaluation}
\label{qualityEvaluation}
\textcolor{black}{To evaluate the quality and reliability of the generated summaries, we focus on two key metrics: Semantic Richness and Hallucination Rate. Semantic Richness is measured by calculating the ratio of L2 and L3 sentences to the total number of sentences in the summaries generated by each model. The Hallucination Rate is determined by the ratio of the number of hallucinations to the total number of sentences. It should be noted that the Hallucination Rate may exceed 1, as a single sentence could contain multiple hallucinations. We conducted a statistical analysis of hallucinations in the summaries generated by GPT-4, VL2NL and ChartInsighter. Each sentence was carefully examined, and every hallucination identified was categorized according to the types of hallucinations outlined in Sec.~\ref{sec:Hallucination Types}.}

\textcolor{black}{Statistical results presented in Tab.~\ref{table_combined} (Quality Evaluation) show that although GPT-4 performs very similarly to our model in terms of semantic richness, its hallucination rate is significantly higher, which means ChartInsighter maintains a better balance between semantic richness and factual accuracy, extracting more data insights and deeper semantic layers, with fewer hallucinations.}

\textcolor{black}{We analyze hallucinations in summaries generated by GPT-4 and found \textcolor{black}{Detail Omission frequently occur.} Additionally, GPT-4 tends to focus on vague information rather than providing specific numerical details. It also often overlooks significant peaks, using terms like ``fluctuate''. The hallucination rate in chart summaries generated by VL2NL is the highest. The most common type of hallucination in VL2NL is also Detail Omission, as its summaries only include L1/L2 content without generating L3 content, leading to missing information and incomplete summaries. VL2NL also frequently makes calculation errors when performing statistical operations (e.g., maxima, minima, medians), resulting in Numerical Value Error. The chart summaries generated by VL2NL also contain many Junk Descriptions, where the content is unrelated to the chart, for example, ``The data is sourced from a file named 1.csv''.} In contrast, ChartInsighter offers precise values and exact time points in its summaries.

\subsection{Algorithm Performance}
\label{Sec:System Complexity}
\begin{figure}[tbp]
    \centering
    \includegraphics[width=1\linewidth]{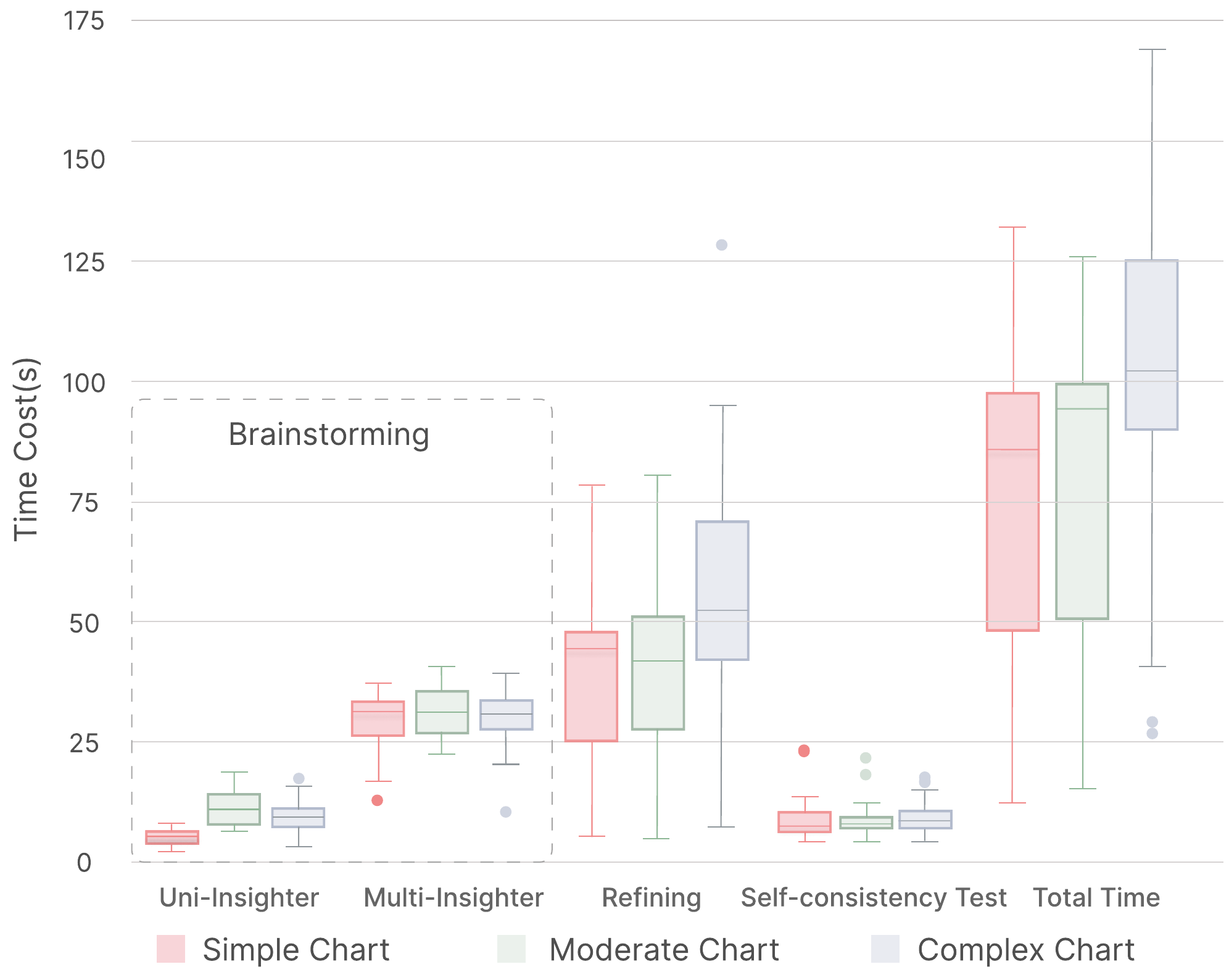}
    \caption{\textcolor{black}{Evaluation results of algorithm performance.} The boxplot displays the time spent on each step processing charts of different complexity—Brainstorming, Refining, and Self-consistency Test—showing the range, median, and outliers for each phase.}
    \label{time}
    \vspace{-8px}
\end{figure}

The algorithm was evaluated on a dataset of 75 charts from our benchmark with varying complexity (25 simple, 25 moderate, and 25 complex).  The tests were conducted by calling the GPT-4 API, and the evaluation was performed at each stage (Fig.~\ref{time}), including Brainstorming, Refining, and Self-consistency Test. The Brainstorming module specifically tested the runtime of two agents, \textit{Uni-Insighter} and \textit{Multi-Insighter}. The observed median execution time for the 75 charts was 87.23 s (± 36.1 s), with a maximum of 171 seconds and a minimum of 18 seconds. As chart complexity increases, the total time also grows.

In Brainstorming, the median time spent by \textit{Uni-Insighter} is 7.6 s ± (4.34 s), balancing accuracy and efficiency. \textit{Multi-Insighter} requires a longer time, 31.4 s ± (5.9 s) because it generates multi-dimensional descriptions three times and conducts majority voting.

Refining is the most time-consuming step, as multiple iterations are inherently time-intensive. However, without this iterative refinement, the accuracy and comprehensiveness of the output would be significantly lower, as evidenced by the summaries generated by GPT-4 and VL2NL in our benchmark. In terms of the median, complex charts take longer than moderate and simple ones, with complex ones requiring 55.34 s (± 23.73 s), moderate 42.31 s (± 20.41 s), and simple 44.01 s ± (19.62 s). Simple and moderate are mostly single-dimensional charts which require less time as they do not involve extracting multi-dimensional insights, while complex charts involve more dimensions and have more intricate trend changes, requiring \textit{Multi-Insighter} to go through more iterations to extract meaningful insights. The average time for the Refining phase is 46.7 s (± 23.7 s), with most charts completed in 2-3 iterations. To prevent inefficient iterations, we set a maximum of 5 iterations, limiting unnecessary time consumption. In the Self-consistency Test phase, the median time is 7.64 s (± 4.34 s), 7.97 s (± 3.78 s), and 9.21 s (± 3.15 s) for simple, moderate, and complex which are faster than Refining since it does not involve multi-dimensional insight extraction.

\subsection{Usage Scenario}
Nancy, a data journalist of a meteorological organization, is working on a report about CO$_{2}$ emissions from coal. She derives historical data on annual CO$_{2}$ emissions from coal in the United Kingdom, the United States, and India, spanning from 1750 to the present, from the website of the Global Carbon Budget Office\cite{globalcarbonbudget}. Using this dataset, she created a multi-dimensional line chart in Vega-Lite specification, visualizing changes in annual coal CO$_{2}$ emissions for these countries.

To quickly extract insights from this complex chart, Nancy uploaded the data table and the Vega-Lite specification to our ChartInsighter system. Then the system generated a chart summary (Fig.~\ref{fig:Interface}-b). The summary covered L1-L3 elements that we proposed in Sec.~\ref{sec:Summary Elements}, including the chart's basic construction, uni-dimensional insights for each of the three countries-upward and downward trends, fluctuations, key extreme values, and comparison between the three countries, highlighting how the U.S. and the U.K. took the lead in emissions in the 19th and 20th centuries, but have significantly decreased in recent decades, and how the developing country, India, has risen to prominence. Nancy agreed with this observation and wishes to include it in her summary.

The summary view was interactive, which can help her quickly map the text to
the relevant portions in the chart, especially in such a complex, multi-dimensional
chart: hovering over \textcolor{black}{the sentence containing data references}  highlighted \textcolor{black}{the corresponding portion} of the chart, allowing Nancy to \textcolor{black}{quickly} verify the accuracy of the summaries. While reviewing, she noticed that using ``sharp'' to describe the U.K. emissions' decline after 1955 is inappropriate compared to the real sharp declines in the U.S. and requested a correction (Fig.~\ref{fig:Interface}-c). After revision, the summary was concise and accurate but missed a key insight—the sharp increase in U.S. emissions in 1944, which she believed had news value and could attract the audience and suggested adding (Fig.~\ref{fig:Interface}-d).
The system updated the summary to include this, delivering a refined version that met her expectations. The final version offered accurate and detailed accounts of the trends for each country, and key turning points—such as peaks in fluctuations in U.S. emissions, the UK’s abrupt shift in its previously steady rise, and comparative analyses of the three countries' emission trends and relations. Nancy used this enhanced summary as the caption for her chart, helping readers easily grasp the insights.
Building on the summary, she crafted a detailed narrative exploring the political and historical context of CO$_{2}$ emissions from coal, enough to be considered a comprehensive analysis.

%% file: sections/7-Discussion.tex
\section{Discussion}
\textcolor{black}{In this section, we discuss some insights gained from our work and potential directions for future research.}

\textbf{Data Types.}  \textcolor{black}{Our research focuses on time-series line charts. We have chosen to study it because it is a fundamental and versatile data type that is prevalent across various fields. Although our current research framework primarily focuses on time-series data, its strong extensibility and generalizability make it suitable for a wider range of data types and chart formats in the future. Future research can build upon our foundation to explore the unique characteristics of other data types in greater depth.}

\textbf{Inputting images.} Research has shown that when generating chart summaries using LLMs, providing backing data leads to more effective summaries compared to other forms of input\cite{kantharaj2022chart,tang2023vistext}. Based on this, we use a combination of raw data and visualization specification as inputs, where visualization specification is used to represent the chart construction. \textcolor{black}{Multimodal Large Language Models have shown strong capability in comprehending images recently\cite{zhang2024tinychart, lin2023sphinx,han2023chartllama,meng2024chartassisstant}. In the future, we can consider including chart images as input, as different types of charts convey rich semantics through unique visual languages, using visual elements such as color, shape, size, and position\cite{huang-etal-2024-chart}, which would enable the model to better infer the characteristics of different chart types.}

\textbf{Hallucinations.} Our work has cataloged the types and frequencies of hallucinations when LLMs generate summaries for time-series data charts, and integrated external modules to address the most frequently occurring hallucinations.  However, only a portion of the hallucinations are mitigated, and other types persist. \textcolor{black}{In the future, Proportion Perception Error can be further improved by applying statistical analysis, such as using Bayesian methods to probabilistically cluster data\cite{HowTallIsTall}, thereby dynamically determining the standard of comparison.
}

\textbf{Integrating domain knowledge.} \textcolor{black}{Our summaries focus on L1-L3 content, which can be detected through visual and data dimensions. Integrating L4 content requires domain-specific knowledge. Additionally, the summary of time-series data charts varies across different fields, each with its own unique characteristics and analytical focus. For example, in the industrial field, the focus is on equipment operating status, fault detection, and potential issues\cite{peifeng2024joint}; whereas in the financial field, it is necessary to account for market fluctuations and the impact of major events on the data. Therefore, generating L4 content in the summary remains a challenging task. One approach is domain-specific fine-tuning of the LLM to improve its understanding and content generation in a particular field. } 

\textbf{Mixed-initiative Summary Generation.} Currently, our interface allows the model to modify the summary but doesn’t provide control over intermediate steps in the generation process. To improve this, we could expose certain functions within the automatic workflow, enabling users to make modifications directly. This would allow finer control, improve hallucination detection, and enhance both system interpretability and user experience, especially for complex \textcolor{black}{chart. To further enhance the flexibility of the interface, future research could extend the existing text-chart link by incorporating a chart-text link as well, where hovering over a region of the chart would highlight the corresponding text, enabling bidirectional interaction.}

\textbf{Time Efficiency.} Although we cannot currently support real-time chart generation, the waiting time of the cold start for generative tasks, as discussed in Sec.~\ref{Sec:System Complexity}, is entirely acceptable, since users can receive more accurate and comprehensive summaries, ultimately saving considerable time that would otherwise be spent on extensive revisions.  One approach is to allow users to view the generation process, during which they can stop the iteration once a satisfactory result is achieved. Another potential future direction is to improve the processing efficiency of large-scale language models by optimizing computational resources and leveraging hardware acceleration\cite{bai2023transformer, menghani2023efficient}.

%% file: sections/8-Conclusion.tex
\section{Conclusion}
We introduce ChartInsighter, an automated system for generating summaries of time-series data charts. We have identified the elements within time-series data chart summaries and, through testing and statistical analysis, we have summarized the types of hallucinations that may occur when LLMs automatically generate summaries. These findings serve as guidelines for our system's generation. We designed a framework that seamlessly integrates natural language inference capabilities with external analysis modules, utilizing multi-agent collaboration for iterative refinement to produce the final chart summary. We constructed a benchmark that annotates hallucinations at a sentence level. We also conducted evaluations to validate our system's capacities. Results confirmed that our system significantly outperforms \textcolor{black}{state-of-the-art} LLMs in generating time-series data chart summaries and effectively mitigates commonly occurring hallucinations. Our guidelines and framework can advance research in automation for chart summary generation.

%% file: main.bbl
\begin{thebibliography}{10}
\renewcommand*{\sfdefault}{PTSansNarrow-TLF}

\bibitem{stlouisfed}
Federal reserve bank of st. louis.
\newblock \url{https://www.stlouisfed.org/}, 2024.
\newblock Accessed: 2024-12-31.

\bibitem{statista}
Statista.
\newblock \url{https://www.statista.com/}, 2024.
\newblock Accessed: 2024-12-31.

\bibitem{achiam2023gpt}
J.~Achiam, S.~Adler, S.~Agarwal, L.~Ahmad, I.~Akkaya, F.~L. Aleman, D.~Almeida, J.~Altenschmidt, S.~Altman, S.~Anadkat, et~al.
\newblock Gpt-4 technical report.
\newblock {\em arXiv preprint arXiv:2303.08774}, 2023.

\bibitem{claude2023}
Anthropic.
\newblock Claude.
\newblock \url{https://www.anthropic.com/claude}, 2023.
\newblock Accessed: 2024-12-31.

\bibitem{bai2023transformer}
Y.~Bai, H.~Zhou, K.~Zhao, J.~Chen, J.~Yu, and K.~Wang.
\newblock Transformer-opu: An fpga-based overlay processor for transformer networks.
\newblock In {\em 2023 IEEE 31st Annual International Symposium on Field-Programmable Custom Computing Machines (FCCM)}, pp. 221--221. IEEE, 2023. \href{https://doi.org/10.1109/FCCM57271.2023.00049}
{doi: \textsf{%
10\hspace{.1pt}\discretionary{.}{%
}{.}\hspace{.4pt}1109\discretionary{/}{%
}{/}FCCM57271\hspace{.1pt}\discretionary{.}{%
}{.}\hspace{.4pt}2023\hspace{.1pt}\discretionary{.}{%
}{.}\hspace{.4pt}00049}}


\bibitem{battle2018beagle}
L.~Battle, P.~Duan, Z.~Miranda, D.~Mukusheva, R.~Chang, and M.~Stonebraker.
\newblock Beagle: Automated extraction and interpretation of visualizations from the web.
\newblock CHI '18,  8 pages, p. 1–8. Association for Computing Machinery, New York, NY, USA, 2018. \href{https://doi.org/10.1145/3173574.3174168}
{doi: \textsf{%
10\hspace{.1pt}\discretionary{.}{%
}{.}\hspace{.4pt}1145\discretionary{/}{%
}{/}3173574\hspace{.1pt}\discretionary{.}{%
}{.}\hspace{.4pt}3174168}}


\bibitem{burns2009modeling}
R.~Burns, S.~Carberry, and S.~Elzer.
\newblock Modeling relative task effort for grouped bar charts.
\newblock In {\em Proceedings of the Annual Meeting of the Cognitive Science Society}, vol.~31, 2009.

\bibitem{chen2023programthoughtspromptingdisentangling}
W.~Chen, X.~Ma, X.~Wang, and W.~W. Cohen.
\newblock Program of thoughts prompting: Disentangling computation from reasoning for numerical reasoning tasks.
\newblock {\em arXiv preprint arXiv:2211.12588}, 2022.

\bibitem{chen2023universal}
X.~Chen, R.~Aksitov, U.~Alon, J.~Ren, K.~Xiao, P.~Yin, S.~Prakash, C.~Sutton, X.~Wang, and D.~Zhou.
\newblock Universal self-consistency for large language model generation.
\newblock {\em arXiv preprint arXiv:2311.17311}, 2023.

\bibitem{cohen2023lm}
R.~Cohen, M.~Hamri, M.~Geva, and A.~Globerson.
\newblock Lm vs lm: Detecting factual errors via cross examination.
\newblock {\em arXiv preprint arXiv:2305.13281}, 2023.

\bibitem{dibia2023lida}
V.~Dibia.
\newblock {LIDA}: A tool for automatic generation of grammar-agnostic visualizations and infographics using large language models.
\newblock In {\em Proceedings of the 61st Annual Meeting of the Association for Computational Linguistics (Volume 3: System Demonstrations)}, pp. 113--126. Association for Computational Linguistics, Toronto, Canada, July 2023. \href{https://doi.org/10.18653/v1/2023.acl-demo.11}
{doi: \textsf{%
10\hspace{.1pt}\discretionary{.}{%
}{.}\hspace{.4pt}18653\discretionary{/}{%
}{/}v1\discretionary{/}{%
}{/}2023\hspace{.1pt}\discretionary{.}{%
}{.}\hspace{.4pt}acl\discretionary{%
}{-}{-}demo\hspace{.1pt}\discretionary{.}{%
}{.}\hspace{.4pt}11}}


\bibitem{llama}
A.~Dubey, A.~Jauhri, A.~Pandey, A.~Kadian, A.~Al-Dahle, A.~Letman, A.~Mathur, A.~Schelten, A.~Yang, A.~Fan, et~al.
\newblock The llama 3 herd of models.
\newblock {\em arXiv preprint arXiv:2407.21783}, 2024.

\bibitem{frieder2024mathematical}
S.~Frieder, L.~Pinchetti, A.~Chevalier, R.-R. Griffiths, T.~Salvatori, T.~Lukasiewicz, P.~Petersen, and J.~Berner.
\newblock Mathematical capabilities of chatgpt.
\newblock In {\em Proceedings of the 37th International Conference on Neural Information Processing Systems}, NIPS '23,  article no. 1205,  46 pages. Curran Associates Inc., Red Hook, NY, USA, 2024. \href{https://dl.acm.org/doi/10.5555/3666122.3667327}
{doi: \textsf{%
doi\discretionary{/}{%
}{/}10\hspace{.1pt}\discretionary{.}{%
}{.}\hspace{.4pt}5555\discretionary{/}{%
}{/}3666122\hspace{.1pt}\discretionary{.}{%
}{.}\hspace{.4pt}3667327}}


\bibitem{fu2011review}
T.-c. Fu.
\newblock A review on time series data mining.
\newblock {\em Engineering Applications of Artificial Intelligence}, 24(1):164--181, 2011. \href{https://doi.org/10.1016/j.engappai.2010.09.007}
{doi: \textsf{%
10\hspace{.1pt}\discretionary{.}{%
}{.}\hspace{.4pt}1016\discretionary{/}{%
}{/}j\hspace{.1pt}\discretionary{.}{%
}{.}\hspace{.4pt}engappai\hspace{.1pt}\discretionary{.}{%
}{.}\hspace{.4pt}2010\hspace{.1pt}\discretionary{.}{%
}{.}\hspace{.4pt}09\hspace{.1pt}\discretionary{.}{%
}{.}\hspace{.4pt}007}}


\bibitem{gao2023chatgpt}
J.~Gao, X.~Ding, B.~Qin, and T.~Liu.
\newblock Is {C}hat{GPT} a good causal reasoner? a comprehensive evaluation.
\newblock In H.~Bouamor, J.~Pino, and K.~Bali, eds., {\em Findings of the Association for Computational Linguistics: EMNLP 2023}, pp. 11111--11126. Association for Computational Linguistics, Singapore, Dec. 2023. \href{https://doi.org/10.18653/v1/2023.findings-emnlp.743}
{doi: \textsf{%
10\hspace{.1pt}\discretionary{.}{%
}{.}\hspace{.4pt}18653\discretionary{/}{%
}{/}v1\discretionary{/}{%
}{/}2023\hspace{.1pt}\discretionary{.}{%
}{.}\hspace{.4pt}findings\discretionary{%
}{-}{-}emnlp\hspace{.1pt}\discretionary{.}{%
}{.}\hspace{.4pt}743}}


\bibitem{greenbacker2011abstractive}
C.~Greenbacker, P.~Wu, S.~Carberry, K.~F. McCoy, and S.~Elzer.
\newblock Abstractive summarization of line graphs from popular media.
\newblock In {\em Proceedings of the Workshop on Automatic Summarization for Different Genres, Media, and Languages}, pp. 41--48, 2011. \href{https://dl.acm.org/doi/10.5555/2018987.2018993}
{doi: \textsf{%
doi\discretionary{/}{%
}{/}10\hspace{.1pt}\discretionary{.}{%
}{.}\hspace{.4pt}5555\discretionary{/}{%
}{/}2018987\hspace{.1pt}\discretionary{.}{%
}{.}\hspace{.4pt}2018993}}


\bibitem{hadi2024large}
M.~U. Hadi, Q.~Al~Tashi, A.~Shah, R.~Qureshi, A.~Muneer, M.~Irfan, A.~Zafar, M.~B. Shaikh, N.~Akhtar, J.~Wu, et~al.
\newblock Large language models: a comprehensive survey of its applications, challenges, limitations, and future prospects.
\newblock {\em Authorea Preprints}, 2024.

\bibitem{han2023chartllama}
Y.~Han, C.~Zhang, X.~Chen, X.~Yang, Z.~Wang, G.~Yu, B.~Fu, and H.~Zhang.
\newblock Chartllama: A multimodal llm for chart understanding and generation.
\newblock {\em arXiv preprint arXiv:2311.16483}, 2023.

\bibitem{he2024leveraging}
Y.~He, S.~Cao, Y.~Shi, Q.~Chen, K.~Xu, and N.~Cao.
\newblock Leveraging large models for crafting narrative visualization: a survey.
\newblock {\em arXiv preprint arXiv:2401.14010}, 2024.

\bibitem{hegarty1993constructing}
M.~Hegarty and M.-A. Just.
\newblock Constructing mental models of machines from text and diagrams.
\newblock {\em Journal of memory and language}, 32(6):717--742, 1993. \href{https://doi.org/10.1006/jmla.1993.1036}
{doi: \textsf{%
10\hspace{.1pt}\discretionary{.}{%
}{.}\hspace{.4pt}1006\discretionary{/}{%
}{/}jmla\hspace{.1pt}\discretionary{.}{%
}{.}\hspace{.4pt}1993\hspace{.1pt}\discretionary{.}{%
}{.}\hspace{.4pt}1036}}


\bibitem{huang2022towards}
J.~Huang and K.~C.-C. Chang.
\newblock Towards reasoning in large language models: A survey.
\newblock {\em arXiv preprint arXiv:2212.10403}, 2022.

\bibitem{huang-etal-2024-chart}
K.-H. Huang, H.~P. Chan, Y.~R. Fung, H.~Qiu, M.~Zhou, S.~Joty, S.-F. Chang, and H.~Ji.
\newblock From pixels to insights: A survey on automatic chart understanding in the era of large foundation models.
\newblock {\em arXiv preprint arXiv:2403.12027}, 2024.

\bibitem{obaid2023tackling}
S.~O.~U. Islam, I.~{\v{S}}krjanec, O.~Du{\v{s}}ek, and V.~Demberg.
\newblock Tackling hallucinations in neural chart summarization.
\newblock pp. 414--423, Sept. 2023. \href{https://doi.org/10.18653/v1/2023.inlg-main.30}
{doi: \textsf{%
10\hspace{.1pt}\discretionary{.}{%
}{.}\hspace{.4pt}18653\discretionary{/}{%
}{/}v1\discretionary{/}{%
}{/}2023\hspace{.1pt}\discretionary{.}{%
}{.}\hspace{.4pt}inlg\discretionary{%
}{-}{-}main\hspace{.1pt}\discretionary{.}{%
}{.}\hspace{.4pt}30}}


\bibitem{kantharaj2022chart}
S.~Kantharaj, R.~T. Leong, X.~Lin, A.~Masry, M.~Thakkar, E.~Hoque, and S.~Joty.
\newblock Chart-to-text: A large-scale benchmark for chart summarization.
\newblock In S.~Muresan, P.~Nakov, and A.~Villavicencio, eds., {\em Proceedings of the 60th Annual Meeting of the Association for Computational Linguistics (Volume 1: Long Papers)}, pp. 4005--4023. Association for Computational Linguistics, Dublin, Ireland, May 2022. \href{https://doi.org/10.18653/v1/2022.acl-long.277}
{doi: \textsf{%
10\hspace{.1pt}\discretionary{.}{%
}{.}\hspace{.4pt}18653\discretionary{/}{%
}{/}v1\discretionary{/}{%
}{/}2022\hspace{.1pt}\discretionary{.}{%
}{.}\hspace{.4pt}acl\discretionary{%
}{-}{-}long\hspace{.1pt}\discretionary{.}{%
}{.}\hspace{.4pt}277}}


\bibitem{kim2023emphasischecker}
D.~H. Kim, S.~Choi, J.~Kim, V.~Setlur, and M.~Agrawala.
\newblock Emphasischecker: A tool for guiding chart and caption emphasis.
\newblock {\em IEEE Transactions on Visualization and Computer Graphics}, 2023. \href{https://doi.org/10.1109/TVCG.2023.3327150}
{doi: \textsf{%
10\hspace{.1pt}\discretionary{.}{%
}{.}\hspace{.4pt}1109\discretionary{/}{%
}{/}TVCG\hspace{.1pt}\discretionary{.}{%
}{.}\hspace{.4pt}2023\hspace{.1pt}\discretionary{.}{%
}{.}\hspace{.4pt}3327150}}


\bibitem{kim2021towards}
D.~H. Kim, V.~Setlur, and M.~Agrawala.
\newblock Towards understanding how readers integrate charts and captions: A case study with line charts.
\newblock In {\em Proceedings of the 2021 CHI Conference on Human Factors in Computing Systems}, pp. 1--11, 2021. \href{https://doi.org/10.1145/3411764.3445443}
{doi: \textsf{%
10\hspace{.1pt}\discretionary{.}{%
}{.}\hspace{.4pt}1145\discretionary{/}{%
}{/}3411764\hspace{.1pt}\discretionary{.}{%
}{.}\hspace{.4pt}3445443}}


\bibitem{kim2018multimodal}
E.~Kim and K.~F. McCoy.
\newblock Multimodal deep learning using images and text for information graphic classification.
\newblock In {\em Proceedings of the 20th International ACM SIGACCESS Conference on Computers and Accessibility}, pp. 143--148, 2018. \href{https://doi.org/10.1145/3234695.3236357}
{doi: \textsf{%
10\hspace{.1pt}\discretionary{.}{%
}{.}\hspace{.4pt}1145\discretionary{/}{%
}{/}3234695\hspace{.1pt}\discretionary{.}{%
}{.}\hspace{.4pt}3236357}}


\bibitem{ko2024natural}
H.-K. Ko, H.~Jeon, G.~Park, D.~H. Kim, N.~W. Kim, J.~Kim, and J.~Seo.
\newblock Natural language dataset generation framework for visualizations powered by large language models.
\newblock In {\em Proceedings of the CHI Conference on Human Factors in Computing Systems}, pp. 1--22, 2024. \href{https://doi.org/10.1145/3613904.3642943}
{doi: \textsf{%
10\hspace{.1pt}\discretionary{.}{%
}{.}\hspace{.4pt}1145\discretionary{/}{%
}{/}3613904\hspace{.1pt}\discretionary{.}{%
}{.}\hspace{.4pt}3642943}}


\bibitem{large1995multimedia}
A.~Large, J.~Beheshti, A.~Breuleux, and A.~Renaud.
\newblock Multimedia and comprehension: The relationship among text, animation, and captions.
\newblock {\em Journal of the American society for information science}, 46(5):340--347, 1995. \href{https://doi.org/10.1002/(SICI)1097-4571(199506)46:5<340::AID-ASI5>3.0.CO;2-S}
{doi: \textsf{%
10\hspace{.1pt}\discretionary{.}{%
}{.}\hspace{.4pt}1002\discretionary{/}{%
}{/}\discretionary{%
}{(}{(}SICI\discretionary{)}{%
}{)}1097\discretionary{%
}{-}{-}4571\discretionary{%
}{(}{(}199506\discretionary{)}{%
}{)}46\discretionary{:}{%
}{:}5{\textless}340\discretionary{:}{%
}{:}\discretionary{:}{%
}{:}AID\discretionary{%
}{-}{-}ASI5{\textgreater}3\hspace{.1pt}\discretionary{.}{%
}{.}\hspace{.4pt}0\hspace{.1pt}\discretionary{.}{%
}{.}\hspace{.4pt}CO\discretionary{;}{%
}{;}2\discretionary{%
}{-}{-}S}}


\bibitem{law2020characterizing}
P.-M. Law, A.~Endert, and J.~Stasko.
\newblock Characterizing automated data insights.
\newblock In {\em 2020 IEEE Visualization Conference (VIS)}, pp. 171--175. IEEE, 2020. \href{https://doi.org/10.1109/VIS47514.2020.00041}
{doi: \textsf{%
10\hspace{.1pt}\discretionary{.}{%
}{.}\hspace{.4pt}1109\discretionary{/}{%
}{/}VIS47514\hspace{.1pt}\discretionary{.}{%
}{.}\hspace{.4pt}2020\hspace{.1pt}\discretionary{.}{%
}{.}\hspace{.4pt}00041}}


\bibitem{li2021kg4vis}
H.~Li, Y.~Wang, S.~Zhang, Y.~Song, and H.~Qu.
\newblock Kg4vis: A knowledge graph-based approach for visualization recommendation.
\newblock {\em IEEE Transactions on Visualization and Computer Graphics}, 28(1):195--205, 2021. \href{https://doi.org/10.1109/TVCG.2021.3114863}
{doi: \textsf{%
10\hspace{.1pt}\discretionary{.}{%
}{.}\hspace{.4pt}1109\discretionary{/}{%
}{/}TVCG\hspace{.1pt}\discretionary{.}{%
}{.}\hspace{.4pt}2021\hspace{.1pt}\discretionary{.}{%
}{.}\hspace{.4pt}3114863}}


\bibitem{lin2023sphinx}
Z.~Lin, C.~Liu, R.~Zhang, P.~Gao, L.~Qiu, H.~Xiao, H.~Qiu, C.~Lin, W.~Shao, K.~Chen, et~al.
\newblock Sphinx: The joint mixing of weights, tasks, and visual embeddings for multi-modal large language models.
\newblock {\em arXiv preprint arXiv:2311.07575}, 2023.

\bibitem{liu2024chartthinker}
M.~Liu, D.~Chen, Y.~Li, G.~Fang, and Y.~Shen.
\newblock Chartthinker: A contextual chain-of-thought approach to optimized chart summarization.
\newblock {\em arXiv preprint arXiv:2403.11236}, 2024.

\bibitem{lundgard2021accessible}
A.~Lundgard and A.~Satyanarayan.
\newblock Accessible visualization via natural language descriptions: A four-level model of semantic content.
\newblock {\em IEEE Transactions on Visualization and Computer Graphics}, 28(1):1073--1083, 2022. \href{https://doi.org/10.1109/TVCG.2021.3114770}
{doi: \textsf{%
10\hspace{.1pt}\discretionary{.}{%
}{.}\hspace{.4pt}1109\discretionary{/}{%
}{/}TVCG\hspace{.1pt}\discretionary{.}{%
}{.}\hspace{.4pt}2021\hspace{.1pt}\discretionary{.}{%
}{.}\hspace{.4pt}3114770}}


\bibitem{ma2021metainsight}
P.~Ma, R.~Ding, S.~Han, and D.~Zhang.
\newblock Metainsight: Automatic discovery of structured knowledge for exploratory data analysis.
\newblock In {\em Proceedings of the 2021 international conference on management of data}, pp. 1262--1274, 2021. \href{https://doi.org/10.1145/3448016.3457267}
{doi: \textsf{%
10\hspace{.1pt}\discretionary{.}{%
}{.}\hspace{.4pt}1145\discretionary{/}{%
}{/}3448016\hspace{.1pt}\discretionary{.}{%
}{.}\hspace{.4pt}3457267}}


\bibitem{masry2023unichart}
A.~Masry, P.~Kavehzadeh, X.~L. Do, E.~Hoque, and S.~Joty.
\newblock Unichart: A universal vision-language pretrained model for chart comprehension and reasoning.
\newblock {\em arXiv preprint arXiv:2305.14761}, 2023.

\bibitem{meng2024chartassisstant}
F.~Meng, W.~Shao, Q.~Lu, P.~Gao, K.~Zhang, Y.~Qiao, and P.~Luo.
\newblock Chartassisstant: A universal chart multimodal language model via chart-to-table pre-training and multitask instruction tuning.
\newblock {\em arXiv preprint arXiv:2401.02384}, 2024.

\bibitem{menghani2023efficient}
G.~Menghani.
\newblock Efficient deep learning: A survey on making deep learning models smaller, faster, and better.
\newblock {\em ACM Computing Surveys}, 55(12):1--37, 2023. \href{https://doi.org/10.1145/3578938}
{doi: \textsf{%
10\hspace{.1pt}\discretionary{.}{%
}{.}\hspace{.4pt}1145\discretionary{/}{%
}{/}3578938}}


\bibitem{narechania2020nl4dv}
A.~Narechania, A.~Srinivasan, and J.~Stasko.
\newblock Nl4dv: A toolkit for generating analytic specifications for data visualization from natural language queries.
\newblock {\em IEEE Transactions on Visualization and Computer Graphics}, 27(2):369--379, 2021. \href{https://doi.org/10.1109/TVCG.2020.3030378}
{doi: \textsf{%
10\hspace{.1pt}\discretionary{.}{%
}{.}\hspace{.4pt}1109\discretionary{/}{%
}{/}TVCG\hspace{.1pt}\discretionary{.}{%
}{.}\hspace{.4pt}2020\hspace{.1pt}\discretionary{.}{%
}{.}\hspace{.4pt}3030378}}


\bibitem{obeid2020chart}
J.~Obeid and E.~Hoque.
\newblock Chart-to-text: Generating natural language descriptions for charts by adapting the transformer model.
\newblock {\em arXiv preprint arXiv:2010.09142}, 2020.

\bibitem{openai_gpt4_2024}
{OpenAI}.
\newblock Hello gpt-4o.
\newblock \url{https://openai.com/index/hello-gpt-4o/}, 2024.
\newblock Accessed: 2024-12-31.

\bibitem{peifeng2024joint}
L.~Peifeng, L.~Qian, X.~Zhao, and B.~Tao.
\newblock Joint knowledge graph and large language model for fault diagnosis and its application in aviation assembly.
\newblock {\em IEEE Transactions on Industrial Informatics}, 2024. \href{https://doi.org/10.1109/TII.2024.3366977}
{doi: \textsf{%
10\hspace{.1pt}\discretionary{.}{%
}{.}\hspace{.4pt}1109\discretionary{/}{%
}{/}TII\hspace{.1pt}\discretionary{.}{%
}{.}\hspace{.4pt}2024\hspace{.1pt}\discretionary{.}{%
}{.}\hspace{.4pt}3366977}}


\bibitem{pewresearch}
{Pew Research Center}.
\newblock Pew research center.
\newblock \url{https://www.pewresearch.org/}, 2024.
\newblock Accessed: 2024-12-31.

\bibitem{globalcarbonbudget}
G.~C. Project.
\newblock Global carbon budget.
\newblock \url{https://globalcarbonbudget.org/}, 2024.
\newblock Accessed: 2024-12-31.

\bibitem{ourworldindata}
M.~Roser, H.~Ritchie, and E.~Ortiz-Ospina.
\newblock Our world in data.
\newblock \url{https://ourworldindata.org/}, 2024.
\newblock Accessed: 2024-12-31.

\bibitem{satyanarayan2016vega}
A.~Satyanarayan, D.~Moritz, K.~Wongsuphasawat, and J.~Heer.
\newblock Vega-lite: A grammar of interactive graphics.
\newblock {\em IEEE Transactions on Visualization and Computer Graphics}, 23(1):341--350, 2016. \href{https://doi.org/10.1109/TVCG.2016.2599030}
{doi: \textsf{%
10\hspace{.1pt}\discretionary{.}{%
}{.}\hspace{.4pt}1109\discretionary{/}{%
}{/}TVCG\hspace{.1pt}\discretionary{.}{%
}{.}\hspace{.4pt}2016\hspace{.1pt}\discretionary{.}{%
}{.}\hspace{.4pt}2599030}}


\bibitem{HowTallIsTall}
L.~Schmidt, N.~Goodman, D.~Barner, B.~Edu, and J.~Tenenbaum.
\newblock How tall is tall? compositionality, statistics, and gradable adjectives.
\newblock 09 2010. \href{https://escholarship.org/uc/item/8ft3x4c7}
{doi: \textsf{%
uc\discretionary{/}{%
}{/}item\discretionary{/}{%
}{/}8ft3x4c7}}


\bibitem{setlur2019inferencing}
V.~Setlur, M.~Tory, and A.~Djalali.
\newblock Inferencing underspecified natural language utterances in visual analysis.
\newblock In {\em Proceedings of the 24th international conference on intelligent user interfaces}, pp. 40--51, 2019. \href{https://doi.org/10.1145/3301275.3302270}
{doi: \textsf{%
10\hspace{.1pt}\discretionary{.}{%
}{.}\hspace{.4pt}1145\discretionary{/}{%
}{/}3301275\hspace{.1pt}\discretionary{.}{%
}{.}\hspace{.4pt}3302270}}


\bibitem{shi2023supporting}
Y.~Shi, B.~Chen, Y.~Chen, Z.~Jin, K.~Xu, X.~Jiao, T.~Gao, and N.~Cao.
\newblock Supporting guided exploratory visual analysis on time series data with reinforcement learning.
\newblock {\em IEEE Transactions on Visualization and Computer Graphics}, 2023. \href{https://doi.org/10.1109/TVCG.2023.3327200}
{doi: \textsf{%
10\hspace{.1pt}\discretionary{.}{%
}{.}\hspace{.4pt}1109\discretionary{/}{%
}{/}TVCG\hspace{.1pt}\discretionary{.}{%
}{.}\hspace{.4pt}2023\hspace{.1pt}\discretionary{.}{%
}{.}\hspace{.4pt}3327200}}


\bibitem{singh2018big}
S.~Singh and A.~Yassine.
\newblock Big data mining of energy time series for behavioral analytics and energy consumption forecasting.
\newblock {\em Energies}, 11(2):452, 2018. \href{https://doi.org/10.3390/en11020452}
{doi: \textsf{%
10\hspace{.1pt}\discretionary{.}{%
}{.}\hspace{.4pt}3390\discretionary{/}{%
}{/}en11020452}}


\bibitem{datatales}
N.~Sultanum and A.~Srinivasan.
\newblock Datatales: Investigating the use of large language models for authoring data-driven articles.
\newblock In {\em 2023 IEEE Visualization and Visual Analytics (VIS)}, pp. 231--235. IEEE, 2023. \href{https://doi.org/10.1109/VIS54172.2023.00055}
{doi: \textsf{%
10\hspace{.1pt}\discretionary{.}{%
}{.}\hspace{.4pt}1109\discretionary{/}{%
}{/}VIS54172\hspace{.1pt}\discretionary{.}{%
}{.}\hspace{.4pt}2023\hspace{.1pt}\discretionary{.}{%
}{.}\hspace{.4pt}00055}}


\bibitem{tang2023vistext}
B.~J. Tang, A.~Boggust, and A.~Satyanarayan.
\newblock {VisText: A Benchmark for Semantically Rich Chart Captioning}.
\newblock In {\em The Annual Meeting of the Association for Computational Linguistics (ACL)}, 2023.

\bibitem{tian2024chartgpt}
Y.~Tian, W.~Cui, D.~Deng, X.~Yi, Y.~Yang, H.~Zhang, and Y.~Wu.
\newblock Chartgpt: Leveraging llms to generate charts from abstract natural language.
\newblock {\em IEEE Transactions on Visualization and Computer Graphics}, 2024. \href{https://doi.org/10.1109/TVCG.2024.3368621}
{doi: \textsf{%
10\hspace{.1pt}\discretionary{.}{%
}{.}\hspace{.4pt}1109\discretionary{/}{%
}{/}TVCG\hspace{.1pt}\discretionary{.}{%
}{.}\hspace{.4pt}2024\hspace{.1pt}\discretionary{.}{%
}{.}\hspace{.4pt}3368621}}


\bibitem{vegalite}
{Vega-Altair Developers}.
\newblock Vega-altair: Declarative visualization in python.
\newblock \url{https://altair-viz.github.io/}, 2024.
\newblock Accessed: 2024-12-31.

\bibitem{wang2023llm4vis}
L.~Wang, S.~Zhang, Y.~Wang, E.-P. Lim, and Y.~Wang.
\newblock {LLM}4{V}is: Explainable visualization recommendation using {C}hat{GPT}.
\newblock In M.~Wang and I.~Zitouni, eds., {\em Proceedings of the 2023 Conference on Empirical Methods in Natural Language Processing: Industry Track}, pp. 675--692. Association for Computational Linguistics, Singapore, Dec. 2023. \href{https://doi.org/10.18653/v1/2023.emnlp-industry.64}
{doi: \textsf{%
10\hspace{.1pt}\discretionary{.}{%
}{.}\hspace{.4pt}18653\discretionary{/}{%
}{/}v1\discretionary{/}{%
}{/}2023\hspace{.1pt}\discretionary{.}{%
}{.}\hspace{.4pt}emnlp\discretionary{%
}{-}{-}industry\hspace{.1pt}\discretionary{.}{%
}{.}\hspace{.4pt}64}}


\bibitem{wang2022self}
X.~Wang, J.~Wei, D.~Schuurmans, Q.~Le, E.~Chi, S.~Narang, A.~Chowdhery, and D.~Zhou.
\newblock Self-consistency improves chain of thought reasoning in language models.
\newblock {\em arXiv preprint arXiv:2203.11171}, 2022.

\bibitem{wang2022detecting}
Y.~Wang, M.~Perry, D.~Whitlock, and J.~W. Sutherland.
\newblock Detecting anomalies in time series data from a manufacturing system using recurrent neural networks.
\newblock {\em Journal of Manufacturing Systems}, 62:823--834, 2022. \href{https://doi.org/10.1016/j.jmsy.2020.12.007}
{doi: \textsf{%
10\hspace{.1pt}\discretionary{.}{%
}{.}\hspace{.4pt}1016\discretionary{/}{%
}{/}j\hspace{.1pt}\discretionary{.}{%
}{.}\hspace{.4pt}jmsy\hspace{.1pt}\discretionary{.}{%
}{.}\hspace{.4pt}2020\hspace{.1pt}\discretionary{.}{%
}{.}\hspace{.4pt}12\hspace{.1pt}\discretionary{.}{%
}{.}\hspace{.4pt}007}}


\bibitem{wang2023unleashing}
Z.~Wang, S.~Mao, W.~Wu, T.~Ge, F.~Wei, and H.~Ji.
\newblock Unleashing the emergent cognitive synergy in large language models: A task-solving agent through multi-persona self-collaboration.
\newblock In K.~Duh, H.~Gomez, and S.~Bethard, eds., {\em Proceedings of the 2024 Conference of the North American Chapter of the Association for Computational Linguistics: Human Language Technologies (Volume 1: Long Papers)}, pp. 257--279. Association for Computational Linguistics, Mexico City, Mexico, June 2024. \href{https://doi.org/10.18653/v1/2024.naacl-long.15}
{doi: \textsf{%
10\hspace{.1pt}\discretionary{.}{%
}{.}\hspace{.4pt}18653\discretionary{/}{%
}{/}v1\discretionary{/}{%
}{/}2024\hspace{.1pt}\discretionary{.}{%
}{.}\hspace{.4pt}naacl\discretionary{%
}{-}{-}long\hspace{.1pt}\discretionary{.}{%
}{.}\hspace{.4pt}15}}


\bibitem{wei2022chain}
J.~Wei, X.~Wang, D.~Schuurmans, M.~Bosma, F.~Xia, E.~Chi, Q.~V. Le, D.~Zhou, et~al.
\newblock Chain-of-thought prompting elicits reasoning in large language models.
\newblock {\em Advances in neural information processing systems}, 35:24824--24837, 2022. \href{https://dl.acm.org/doi/10.5555/3600270.3602070}
{doi: \textsf{%
doi\discretionary{/}{%
}{/}10\hspace{.1pt}\discretionary{.}{%
}{.}\hspace{.4pt}5555\discretionary{/}{%
}{/}3600270\hspace{.1pt}\discretionary{.}{%
}{.}\hspace{.4pt}3602070}}


\bibitem{xia2024chartx}
R.~Xia, B.~Zhang, H.~Ye, X.~Yan, Q.~Liu, H.~Zhou, Z.~Chen, M.~Dou, B.~Shi, J.~Yan, et~al.
\newblock Chartx \& chartvlm: A versatile benchmark and foundation model for complicated chart reasoning.
\newblock {\em arXiv preprint arXiv:2402.12185}, 2024.

\bibitem{xu2024exploring}
Z.~Xu and E.~Wall.
\newblock { Exploring the Capability of LLMs in Performing Low-Level Visual Analytic Tasks on SVG Data Visualizations }.
\newblock In {\em 2024 IEEE Visualization and Visual Analytics (VIS)}, pp. 126--130. IEEE Computer Society, Los Alamitos, CA, USA, Oct. 2024. \href{https://doi.org/10.1109/VIS55277.2024.00033}
{doi: \textsf{%
10\hspace{.1pt}\discretionary{.}{%
}{.}\hspace{.4pt}1109\discretionary{/}{%
}{/}VIS55277\hspace{.1pt}\discretionary{.}{%
}{.}\hspace{.4pt}2024\hspace{.1pt}\discretionary{.}{%
}{.}\hspace{.4pt}00033}}


\bibitem{yang2024foundation}
W.~Yang, M.~Liu, Z.~Wang, and S.~Liu.
\newblock Foundation models meet visualizations: Challenges and opportunities.
\newblock {\em Computational Visual Media}, pp. 1--26, 2024. \href{https://doi.org/10.1007/s41095-023-0393-x}
{doi: \textsf{%
10\hspace{.1pt}\discretionary{.}{%
}{.}\hspace{.4pt}1007\discretionary{/}{%
}{/}s41095\discretionary{%
}{-}{-}023\discretionary{%
}{-}{-}0393\discretionary{%
}{-}{-}x}}


\bibitem{ye2024generative}
Y.~Ye, J.~Hao, Y.~Hou, Z.~Wang, S.~Xiao, Y.~Luo, and W.~Zeng.
\newblock Generative ai for visualization: State of the art and future directions.
\newblock {\em Visual Informatics}, 8(2):43--66, 2024. \href{https://doi.org/10.1016/j.visinf.2024.04.003}
{doi: \textsf{%
10\hspace{.1pt}\discretionary{.}{%
}{.}\hspace{.4pt}1016\discretionary{/}{%
}{/}j\hspace{.1pt}\discretionary{.}{%
}{.}\hspace{.4pt}visinf\hspace{.1pt}\discretionary{.}{%
}{.}\hspace{.4pt}2024\hspace{.1pt}\discretionary{.}{%
}{.}\hspace{.4pt}04\hspace{.1pt}\discretionary{.}{%
}{.}\hspace{.4pt}003}}


\bibitem{yuan2023well}
Z.~Yuan, H.~Yuan, C.~Tan, W.~Wang, and S.~Huang.
\newblock How well do large language models perform in arithmetic tasks?
\newblock {\em arXiv preprint arXiv:2304.02015}, 2023.

\bibitem{zhang2024tinychart}
L.~Zhang, A.~Hu, H.~Xu, M.~Yan, Y.~Xu, Q.~Jin, J.~Zhang, and F.~Huang.
\newblock {T}iny{C}hart: Efficient chart understanding with program-of-thoughts learning and visual token merging.
\newblock In Y.~Al-Onaizan, M.~Bansal, and Y.-N. Chen, eds., {\em Proceedings of the 2024 Conference on Empirical Methods in Natural Language Processing}, pp. 1882--1898. Association for Computational Linguistics, Miami, Florida, USA, Nov. 2024. \href{https://doi.org/10.18653/v1/2024.emnlp-main.112}
{doi: \textsf{%
10\hspace{.1pt}\discretionary{.}{%
}{.}\hspace{.4pt}18653\discretionary{/}{%
}{/}v1\discretionary{/}{%
}{/}2024\hspace{.1pt}\discretionary{.}{%
}{.}\hspace{.4pt}emnlp\discretionary{%
}{-}{-}main\hspace{.1pt}\discretionary{.}{%
}{.}\hspace{.4pt}112}}


\bibitem{zhang2023adavis}
S.~Zhang, H.~Li, H.~Qu, and Y.~Wang.
\newblock Adavis: Adaptive and explainable visualization recommendation for tabular data.
\newblock {\em IEEE Transactions on Visualization and Computer Graphics}, 30(9):5923--5938, 2023. \href{https://doi.org/10.1109/TVCG.2023.3316469}
{doi: \textsf{%
10\hspace{.1pt}\discretionary{.}{%
}{.}\hspace{.4pt}1109\discretionary{/}{%
}{/}TVCG\hspace{.1pt}\discretionary{.}{%
}{.}\hspace{.4pt}2023\hspace{.1pt}\discretionary{.}{%
}{.}\hspace{.4pt}3316469}}


\bibitem{zhang2024self}
W.~Zhang, Y.~Shen, L.~Wu, Q.~Peng, J.~Wang, Y.~Zhuang, and W.~Lu.
\newblock Self-contrast: Better reflection through inconsistent solving perspectives.
\newblock In L.-W. Ku, A.~Martins, and V.~Srikumar, eds., {\em Proceedings of the 62nd Annual Meeting of the Association for Computational Linguistics (Volume 1: Long Papers)}, pp. 3602--3622. Association for Computational Linguistics, Bangkok, Thailand, Aug. 2024. \href{https://doi.org/10.18653/v1/2024.acl-long.197}
{doi: \textsf{%
10\hspace{.1pt}\discretionary{.}{%
}{.}\hspace{.4pt}18653\discretionary{/}{%
}{/}v1\discretionary{/}{%
}{/}2024\hspace{.1pt}\discretionary{.}{%
}{.}\hspace{.4pt}acl\discretionary{%
}{-}{-}long\hspace{.1pt}\discretionary{.}{%
}{.}\hspace{.4pt}197}}


\bibitem{lightva}
Y.~Zhao, J.~Wang, L.~Xiang, X.~Zhang, Z.~Guo, C.~Turkay, Y.~Zhang, and S.~Chen.
\newblock Lightva: Lightweight visual analytics with llm agent-based task planning and execution.
\newblock {\em IEEE Transactions on Visualization and Computer Graphics}, pp. 1--13, 2024. \href{https://doi.org/10.1109/TVCG.2024.3496112}
{doi: \textsf{%
10\hspace{.1pt}\discretionary{.}{%
}{.}\hspace{.4pt}1109\discretionary{/}{%
}{/}TVCG\hspace{.1pt}\discretionary{.}{%
}{.}\hspace{.4pt}2024\hspace{.1pt}\discretionary{.}{%
}{.}\hspace{.4pt}3496112}}


\bibitem{leva}
Y.~Zhao, Y.~Zhang, Y.~Zhang, X.~Zhao, J.~Wang, Z.~Shao, C.~Turkay, and S.~Chen.
\newblock Leva: Using large language models to enhance visual analytics.
\newblock {\em IEEE Transactions on Visualization and Computer Graphics}, pp. 1--17, 2024. \href{https://doi.org/10.1109/TVCG.2024.3368060}
{doi: \textsf{%
10\hspace{.1pt}\discretionary{.}{%
}{.}\hspace{.4pt}1109\discretionary{/}{%
}{/}TVCG\hspace{.1pt}\discretionary{.}{%
}{.}\hspace{.4pt}2024\hspace{.1pt}\discretionary{.}{%
}{.}\hspace{.4pt}3368060}}


\bibitem{zheng2023does}
S.~Zheng, J.~Huang, and K.~C.-C. Chang.
\newblock Why does chatgpt fall short in answering questions faithfully.
\newblock {\em arXiv preprint arXiv:2304.10513}, 2023.

\end{thebibliography}
